\documentclass[letterpaper]{article} 
\usepackage[submission]{aaai25}  
\usepackage{times}  
\usepackage{helvet}  
\usepackage{courier}  
\usepackage[hyphens]{url}  
\usepackage{graphicx} 
\usepackage{booktabs} 
\usepackage{tikz}
\usepackage{helvet}
\usepackage{courier}
\usepackage{amsmath}
\usepackage{amssymb}
\usepackage{subfigure}
\usepackage{pgfplots}
\usepackage{subfig}
\usepgfplotslibrary{fillbetween}
\pgfplotsset{compat=1.18} 

\usepackage{xcolor}
\usepackage{natbib}  
\usepackage{caption} 
\usepackage{algorithm}
\usepackage{algorithmic}

\usepackage{newfloat}
\usepackage{listings}
\DeclareCaptionStyle{ruled}{labelfont=normalfont,labelsep=colon,strut=off} 
\floatstyle{ruled}
\newfloat{listing}{tb}{lst}{}
\floatname{listing}{Listing}
\definecolor{darkgreen}{RGB}{0, 100, 0} 

\begin{document}

\title{On the Benefits of Instance Decomposition in Video Prediction Models}
\author{
    Eliyas Suleyman\textsuperscript{\rm 1}, Paul Henderson\textsuperscript{\rm 1}, Nicolas Pugeault\textsuperscript{\rm 1} 
}
\affiliations{
    \textsuperscript{\rm 1}School of Computing Science, University of Glasgow\\


    2683522s@student.gla.ac.uk, \{paul.henderson,nicolas.pugeault\}@glasgow.ac.uk
%
}


\maketitle

\begin{abstract}
Video prediction is a crucial task for intelligent agents such as robots and autonomous vehicles, since it enables them to anticipate and act early on time-critical incidents. State-of-the-art video prediction methods typically model the dynamics of a scene jointly and implicitly, without any explicit decomposition into separate objects. This is challenging and potentially sub-optimal, as every object in a dynamic scene has their own pattern of movement, typically somewhat independent of others. In this paper, we investigate the benefit of explicitly modeling the objects in a dynamic scene separately within the context of latent-transformer video prediction models. We conduct detailed and carefully-controlled experiments on both synthetic and real-world datasets; our results show that decomposing a dynamic scene leads to higher quality predictions compared with models of a similar capacity that lack such decomposition.
\end{abstract}

\section{Introduction}
Video prediction is the task of predicting future frames based on past frames; it has many applications including autonomous driving \cite{Yang_2024_CVPR}, weather forecasting from satellite images \cite{nature-weather}, and even building general world models \cite{wang2024worlddreamer}.
Predicting future frames 
is challenging, since these are high-dimensional and result from multiple objects' appearances, dynamics and mutual interactions. For example, consider the environment observed while driving a car.
To accurately predict the future, we must identify all objects in our field of vision and estimate their likely movements.
Different types of objects (e.g.~cars, pedestrians, dogs) have very different appearances, but also diverse patterns of movement, and may exhibit complex interactions with other objects.

To reduce this complexity, a natural approach to video prediction is to decompose the scene into several parts \cite{moso, SawareVP, ReHire, DDPAE}.
This enables modeling the appearance and dynamics of each part separately during prediction, thus reducing computational cost and increasing statistical efficiency.
Several works have achieved promising results by such approaches, using different choices of decomposition.
For example, \cite{DDPAE} uses DRNet \cite{DRNET} to learn a disentangled representation of appearance and 2D pose, while \cite{SawareVP, ReHire} use semantic segmentation models, and \cite{moso} separates the foreground, motion and background.
\cite{wu2022slotformer} uses object-centric representation learning \cite{locatello2020object} to separate objects without supervision, and model the dynamics with a multi-slot transformer.

While these works on object-decomposed prediction often achieve impressive results, they do not typically focus on measuring the benefit of object decomposition in a scientifically-controlled way, i.e.~keeping confounding factors such as the number of network parameters, architecture or latent dimensionality constant. 
Moreover, most of these works did not use the modern large latent-space Transformer architectures \cite{vaswani2017attention} that now yield excellent results on diverse domains of videos \cite{yan2021videogpt,wu2024ivideogpt}; they instead used older, smaller CNN- or RNN-based models.

In this work, we perform a detailed study of the benefits of explicitly modeling different objects separately during video prediction, when using modern latent transformer models.
Rather than introducing an entirely new model, we develop a family of architectures that uses ideas from VideoGPT, MOSO and Slotformer~\cite{yan2021videogpt, moso, wu2022slotformer}, 
but supports both monolithic and object-decomposed prediction in a unified framework.
This allows us to perform controlled experiments on the benefits of object decomposition and strategies for modeling interactions.
Specifically, we adopt a hierarchical approach that explicitly decomposes a dynamic scene into individual objects using an instance segmentation model, before encoding these into separate latent spaces.
Going beyond previous approaches, we also mitigate the inefficiency of having separate network parameters per object instance\cite{relational-attention} by sharing parameters across all instances of each class.

We find that even with large transformers, object decomposition leads to considerable improvements in handling complex scenes with multiple interacting objects compared to non-object-centric predictors with similar parameter counts and latent dimensions.

Our main contributions are as follows:
\begin{itemize}

    \item We present the first systematic and comprehensive analysis of the benefits of explicit object decomposition for latent transformer video prediction models.

    \item To achieve this, we develop a scalable framework for video prediction that supports both the object-centric and nondecomposed settings.  
    
    \item We mitigate inefficiencies in object-centric video predictors by sharing weights (and thus knowledge about object dynamics) across slots within each object class.

\end{itemize}

\section{Related Work} \label{relatedwork}

\subsubsection{Recurrent models for video prediction}
Early video prediction models were typically based on the combination of Convolutional Neural Networks \cite{CNN} and Recurrent Neural Networks, often~LSTMs \cite{CONVLSTM, PredRNN, PredRNN++, STAM, SimVP, SGV}. \cite{ReHire} proposed a method to predict future semantic maps, then used those predicted maps to formulate the actual future frames. \cite{SawareVP} proposed a similar approach which decomposes the scene by semantic map then use separate pathways to model the dynamics of different semantic classes.
Of these, some methods are deterministic, i.e.~make a single most-likely prediction of the future \cite{CONVLSTM,PredRNN++}, while others are stochastic, i.e.~sample an autoregressive posterior distribution on possible future frames \cite{SGV,ReHire}. We focus on the stochastic setting in this work since it reflects the fact that the future is inherently uncertain, as well as typically producing sharper predictions.

\subsubsection{Transformer models for video prediction}
 
Following their success on text \cite{vaswani2017attention} and images \cite{Vit}, Transformers have also been applied to video prediction.
A common approach is to first use an encoder network to map the original video frames into a sequence of lower-dimensional latent vectors.
Most models use VQ-VAE \cite{VQVAE} or VQ-GAN \cite{VQGAN} as their encoding network due to their high fidelity reconstruction of original frames, and discrete latent space that enables treating the latents similarly to text tokens.
\cite{yan2021videogpt} proposed the first autoregressive video prediction model based on VQ-GAN and a decoder transformer to predict future frames; iVideoGPT \cite{wu2024ivideogpt} improves performance further. \cite{gupta2022maskvit} proposed a similar method that uses VQ-VAE and transformer, but trains with iterative masking to let it gradually capture the motion patterns in a video. \cite{moso} proposed a pipeline that decomposes the dynamic scene into motion, object and background, then uses a stochastic transformer to predict future frames in latent space. 
Our work also uses a latent transformer, but with an explicit decomposition of the latent space into separate objects, and cross-attention to capture object interactions.

\subsubsection{Diffusion models for video prediction}
The invention of diffusion models \cite{pmlr-v37-sohl-dickstein15,DDPM} and the computationally faster latent diffusion \cite{diffusion} brought significant improvement on many generative tasks. Latent diffusion was originally designed to generate high-resolution images, but has now been applied to video \cite{blattmann2023videoldm,blattmann2023stable,sora}. \cite{videodiff} use a diffusion model to generate long videos via a joint training paradigm with conditional sampling. \cite{hoppe2022diffusion} use a slightly different training process that instead of adding noise to the entire video, randomly kept some of the input frames without noise. \cite{projecteddiff} proposed an interesting way of modelling latent vectors in three different direction by slicing 3D feature vectors along different axes. SORA \cite{sora} is the state-of-the-art video generation model which can generate extremely realistic videos by using diffusion with a transformer architecture. It is able to accurately model complex interactions involving multiple objects \cite{liu2024sora}.

\subsubsection{Object-centric video prediction}
Object-centric representation learning aims to learn decomposed representations of images \cite{locatello2020object,engelcke2019genesis} or videos \cite{Jiang2020SCALOR,zhou2022slot} without supervision.
This can be used to aid video prediction by learning an object-centric predictor (typically a transformer) over the resulting representations \cite{kipf2021conditional, NEURIPS2021_593906af, sajjadi2022object, singh2022simple}.
\cite{relational-attention} use an attention mechanism to learn the relationship between different objects in the video sequence and achieved good results on synthetic CLEVRER \cite{clevrer} dataset. \cite{schmeckpeper2021object} use Mask R-CNN \cite{he2017mask} to get bounding boxes for each entity in the scene, then predict the next state of each bounding box from a single frame. Finally, \cite{henderson20neurips,henderson21corr} proposed self-supervised object-centric approaches that predict frames via latent 3D objects and scene structure from 2D video.

\begin{figure*}[t]
    
    \centering

    \includegraphics[width=0.85\textwidth]{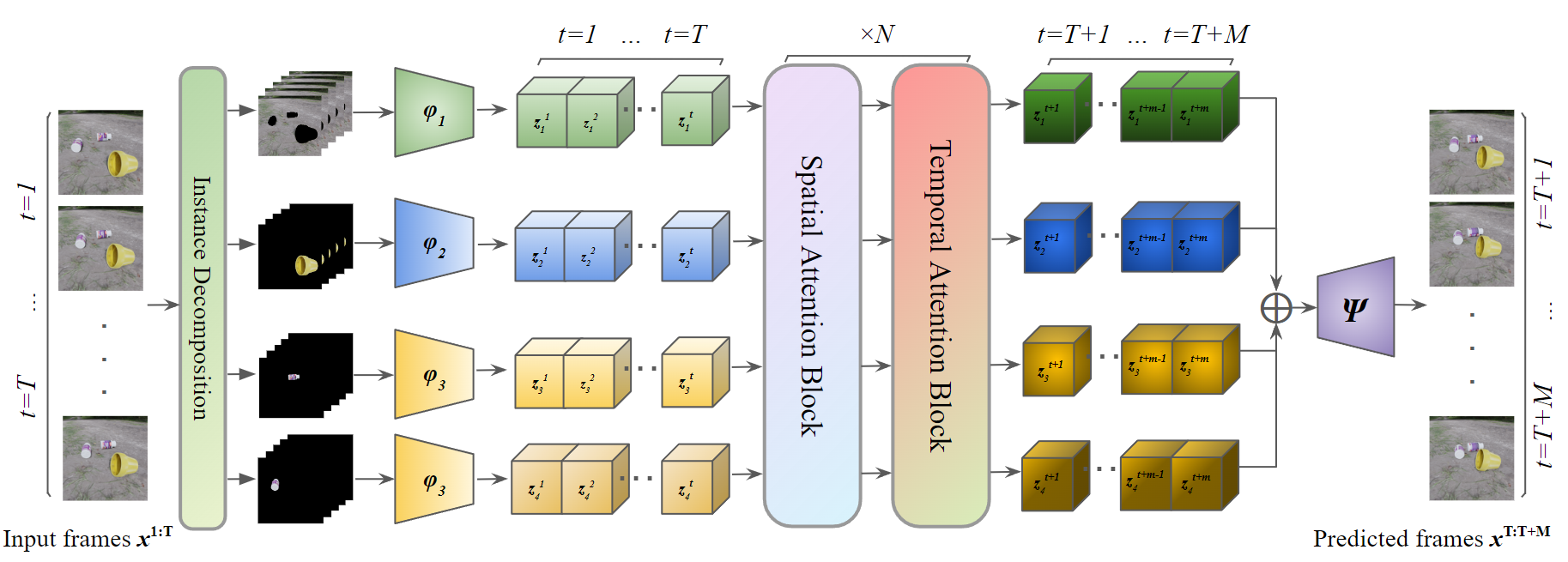} \label{fig:full-model}
    \includegraphics[width=0.85\textwidth]{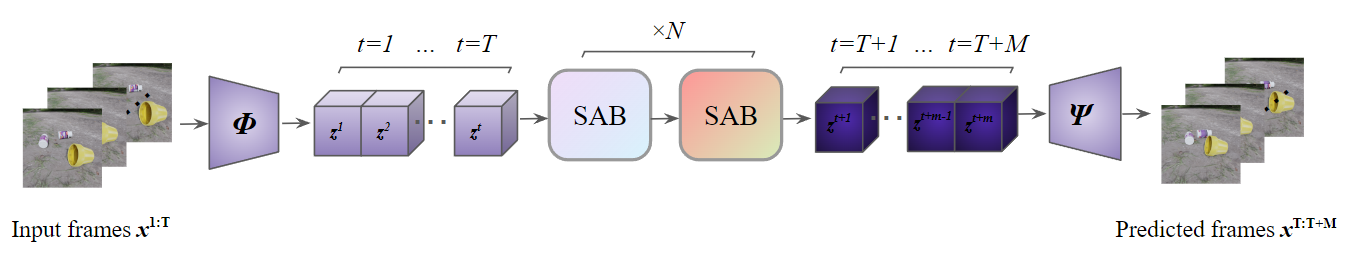} \label{fig:SiS}

    \caption{\textbf{Top:} Our proposed multi-object interacting model \textbf{SCAT}. First the input frames are decomposed via a segmentation model, then each decomposed sequence passes through class-specific encoder to convert the 2D frames into latent representations; then class-specific transformer blocks learn and predict the dynamics of each instance and its relationships with other instances in latent space; lastly, the predicted latent representation are decoded via joint decoder to reconstruct the predicted RGB frames. \textbf{Bottom:} The non-decomposed single-slot variant \textbf{SiS} where the scene is modeled globally and jointly.
    } 
    \vspace{-10pt}
    \centering
    \label{fig:overview}
\end{figure*}

\subsubsection{Cross-attention}
Our model uses cross-attention between instances to capture object interactions.
Similar ideas have been used in many other domains, e.g.~\cite{Zhu_2022_CVPR} use pairwise cross-attention to re-identify pedestrians; \cite{shi2024detail} use cross-attention to fuse information from audio and video for emotion recognition; \cite{CAST} use pairwise cross attention on video action recognition; \cite{diffusion} uses cross attention between image features and text embeddings for conditional image generation.

\section{Methodology} \label{method}

Let $X^{1:T}=\langle x^1,x^2,...,x^T \rangle$, be a sequence of $T$ RGB frames from a video clip, where $x^t\in\mathbb{R}^{h \times w \times 3}$. 
Our goal is to learn a probability distribution on future frames $X^{T+1:T+M}$, conditioned on the preceding frames $X^{1:T}$.

We hypothesize that predicting future frames is more effective when modeling each object or instance separately rather than modeling the entire scene at once.
Moreover, when objects are decomposed, we aim to measure the degree to which cross-attention enables learning interactions among objects, thus making prediction more accurate. 

To test this hypothesis, we design a family of models that support differing degrees of object decomposition and interaction within a unified framework.
We decompose a scene into individual objects using instance segmentation models \cite{yolov8,lueddecke22clipseg}.
The video prediction models then comprise an \textit{object-aware auto-encoder} (OAAE) (Section~\ref{sec:method:segmentation}), which extracts latent representations for each object, and a multi-object transformer (Section~\ref{sec:method:prediction}) that predicts future latent representations conditioned on previous ones; the OAAE is used to decode these future latents back into video frames.
To test our hypotheses, we propose three variants of our overall pipeline:
\begin{itemize}
    
\item  \textbf{Single Slot (SiS)}: Objects are not modeled separately; frames are encoded with a single encoder, and a standard (not object-centric) transformer network is used to predict future frames; this is similar to VideoGPT \cite{yan2021videogpt}.
\item  \textbf{Stochastic non-Class Attended Transformer (SNCAT)}: The scene is decomposed into instances; both the encoder and predictor have one slot for each object in the scene, with parameters shared across instances of the same class, and no interactions among different object slots in the transformer.
\item  \textbf{Stochastic Class Attendted Transformer (SCAT)}: Our full model, which encodes instances separately, then uses a multi-slot transformer for future prediction, with cross-attention to capture object interactions.
\end{itemize} 
The overall pipeline of the fully-interacting decomposed \textbf{SCAT} and single slot \textbf{SiS} models are shown in Figure~\ref{fig:overview}.

\subsection{Object-aware autoencoder}\label{sec:method:segmentation}

We now discuss the encoder we use for extracting the latent representation of a video, which will be used in Section~\ref{sec:method:prediction} as a lower-dimensional space for future prediction.
We first explain the object-aware autoencoder (OAAE) as used in the \textbf{SCAT} and \textbf{SNCAT} models, then give a brief explanation of the simpler (non-object-centric) variant used in \textbf{SiS}.

\subsubsection{Instance decomposition}
Let \( x \in \mathbb{R}^{h \times w \times 3} \) be a frame in an RGB video sequence of width \( w \) and height \( h \). It is decomposed into a set of instances categorized with corresponding class labels using a segmentation model \cite{yolov8,lueddecke22clipseg}. 
After passing the frame to the segmentation model, the segmentor returns a non-overlapping binary mask for each instance $k$, each belonging to one class $c_k\in\{1,\ldots,m\}$; we then multiply the input frame by the respective masks to isolate each object.
The $k$\textsuperscript{th} masked instance is denoted by \( \tilde{x}_k \) for \( k \in \{1, 2, \ldots, N\} \), and its class is denoted as $c_k$.
Assuming the segmentation is panoptic and covers all pixels of the frame, the original frame can be reconstructed by recombining all instances of all classes additively as follows:
\begin{equation}\label{eq:sum_of_semantics}
    x = \sum_{k=1}^{N} {\tilde{x}_{k}}
\end{equation}

\subsubsection{Instance embedding}

We modify the standard VQ-VAE \cite{VQVAE} model to have a set of encoders $\Phi = \{ \phi_1,\phi_2,...,\phi_m \}$ and a set of embedding code books $E = \{ e_1,e_2,...,e_m\}$, each associated with an individual semantic class. Each instance frame $ \tilde{x}_{k}$ is passed to the corresponding encoder $\phi_{c_k}$ and quantized with $e_{c_k}$ to produce a latent vector $\tilde{z}_{k}$:
\begin{equation}\label{eq:encode}
	\tilde{z}_{k} = e_{c_k}^i \mbox{  where  } i = \arg\min_j(\Vert \phi_{c_k}(\tilde{x}_{k}) - e_{c_k}^j\Vert_{2}) 
\end{equation}
The quantized representations are then concatenated into a single vector $z = \bigoplus_{k=1}^{N}\tilde{z}_{k}$ that encodes the complete frame $x$. For convenience, we will use the notation $z = \Phi(x)$ to denote this encoding operation. 
This latent representation $z$ is passed to a single joint decoder $\Psi$ to reconstruct the full frame, i.e.~$\hat{x} = \Psi(z)$. After each up-sampling convolutional layer in the decoder, we incorporate Frequency Complement Modules (FCM) \cite{favae2023cvpr} to learn not only from the target frame but also from feature maps between encoder and decoder.

\subsubsection{Loss function}
Since our OAAE is a multi-object extended version of the original VQ-VAE \cite{VQVAE} with some features of FA-VAE \cite{favae2023cvpr}, we also extend the original loss functions correspondingly.
There are 4 losses: feature loss, commitment loss, vector quantisation loss (VQ loss) and reconstruction loss.
Following \cite{favae2023cvpr}, we impose a loss on feature maps, not only on the final pixels; similar to them we use focal frequency loss (FFL \cite{jiang2021focal}) between the output of encoder convolution layers and decoder FCM layers:
\begin{equation}\label{eq:loss_cvqvae_ffl2}
\mathcal{L}_{feature} = \sum^m_{c=1}\sum^{L-1}_{l=0} \mathit{FFL}(f^c_l, g_{L-l})
\end{equation}
where $c$ indexes encoders (recall there is one per class), $l$ indexes over convolutional layers in the $c$\textsuperscript{th} encoder and $L-l$ over corresponding FCM layers in the decoder ($L$ is the total number of decoder layers). $f_l$ represents the feature map of the $l$\textsuperscript{th} encoder layer, and $g_l$ that of the $l$\textsuperscript{th} FCM module in the decoder.
The VQ and commitment losses are similar to the original VQ-VAE, except we compute these for each class $c$ and instance $k$ then sum over these:
\begin{equation}\label{eq:loss_cvqvae_vq}
\mathcal{L}_{VQ} = \sum_{c=1}^{m}\sum_{k=1}^{n_c}{\Vert \operatorname{sg}[{\tilde{z}^{c}_{k}}] - e_c \Vert_{2}^{2}}
\end{equation}
\begin{equation}\label{eq:loss_cvqvae_commit}
\mathcal{L}_{commitment} = \sum_{c=1}^{m}\sum_{k=1}^{n_c}{\Vert {\tilde{z}^{c}_{k}}-\operatorname{sg}[e_c] \Vert_{2}^{2}}
\end{equation}
where $\operatorname{sg}$ is the stop-gradient operator.
Finally, the reconstruction loss is composed of pixel-space and frequency-space terms calculated between the reconstructed and original frames:
\begin{equation}\label{eq:loss_cvqvae_recon}
\mathcal{L}_{recon} = -\log{p(x|\Psi(\Phi(x)))} + \mathit{FFL}(x, \Psi(\Phi(x)))
\end{equation}
Putting all four terms together yields the final loss function for training OAAE:
\begin{equation}\label{eq:loss_cvqvae}
\mathcal{L}_\mathit{oaae} = \mathcal{L}_{recon} + {\alpha}\mathcal{L}_{feature} + \mathcal{L}_{VQ} + {\beta}\mathcal{L}_{commitment}
\end{equation}
where $\alpha$ and $\beta$ weight the different loss terms.
Once the OAAE is trained, we denote the latent representation for the frame $x^t$ at time step $t$ as $z^t$. This provides a structured and disentangled representation, capturing $N$ instances across $m$ classes.

\subsubsection{Variations of the OAAE} In order to measure whether object decomposition helps with prediction, we also define a non-decomposed version of the VQ-VAE, for use in model \textbf{SiS}.
This only takes the original non-segmented frame as input. It is processed by a single encoder, with the latent size matched to the total latent size (over all instances) for model \textbf{SCAT}. In terms of losses, $L_{recon}$ remains unchanged, $L_{VQ}$, $L_{commitment}$ and $L_{feature}$ will be a modified to a single term without summation since there is now a single encoder and codebook, and feature maps from just one instance.
For the \textbf{SNCAT} model variant, the OAAE is identical to the main version for \textbf{SCAT}, only the subsequent transformer stage is different.

\subsection{Prediction Model} \label{sec:method:prediction}

Using the OAAE, a video clip $X$ is encoded as a sequence of latent representations $Z = \langle z^1, z^2, \ldots, z^T \rangle$. To learn the instance dynamics and its relationship with other instances, we modify the original decoder-only transformer \cite{vaswani2017attention, GPT} into a slot-per-instance auto-regressive transformer that has cross-attention between instances, and shares parameters across instances of each class.

\begin{figure}[t]
    \centering
    \includegraphics[width=0.49\linewidth]{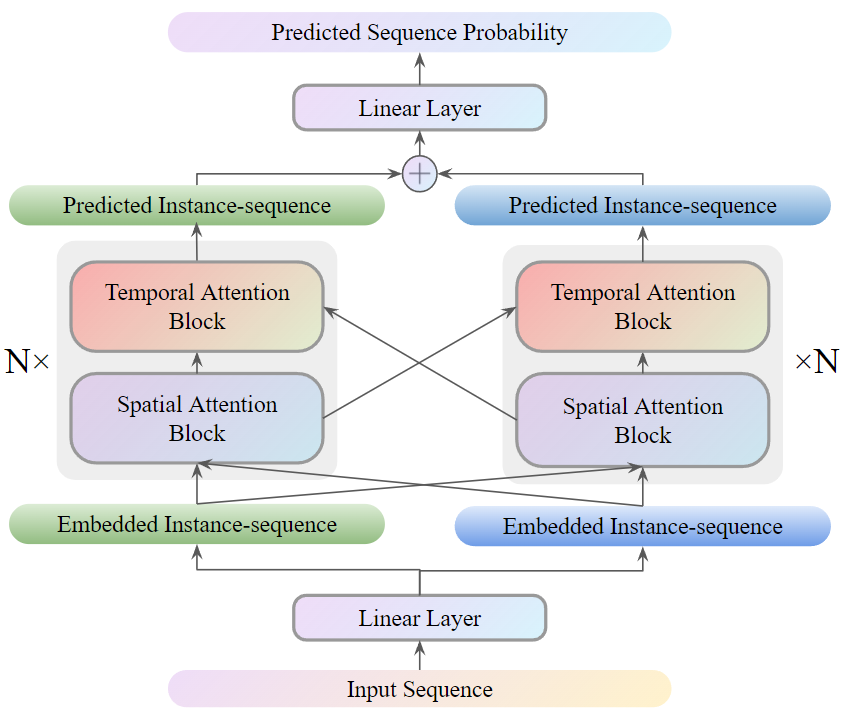}
    \includegraphics[width=0.49\linewidth]{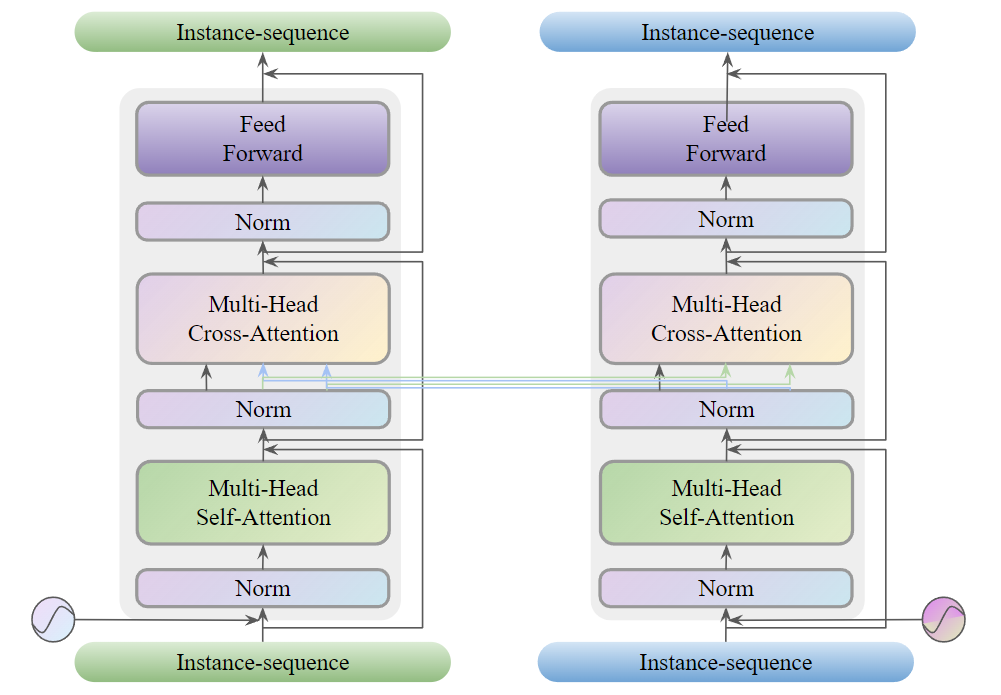}
    \caption{\textbf{Left:} Architecture of the multi-object latent transformer. \textbf{Right:} Detail of spatial and temporal attention blocks. 
    }
    \label{fig:scat}
    \label{fig:block-details}
    \vspace{-10pt}
\end{figure}

Our transformer consists of alternating attention and feed-forward blocks.
However, unlike typical 1D transformers, it includes factored spatial and temporal attention blocks; each of these is applied both for self-attention (i.e.~each instance independently attending to other locations / time-points of itself), and cross-attention (i.e.~each instance attending to different locations / time-points of all other instances).
We use PreNorm \cite{prenorm} in each transformer block.
The output vectors for each instance from the last transformer layer are concatenated and passed through a linear layer. The output size matches the number of embeddings in OAAE, allowing the model to predict the probability of possible indices of future frames.

Because the latent vectors produced by the OAAE are a concatenation of each object instance's latent encoding, we can write the sequence of latent encodings in the video for each individual object instance as $\tilde{Z}_k = \langle z^{1}_{k}, z^{2}_{k},...,z^{T}_{k} \rangle$ where $k$ is the $k_{th}$ instance.

\subsubsection{Spatial and temporal extensions of attention layers} Since an instance latent sequence $\tilde{Z}_{k}$ has a 3-dimensional shape $t \times (h \times w) \times c$, where $c$ represents embedding dimension in OAAE, 
it encompasses both temporal and spatial information. Merely flattening the latent vector to form the video sequence in latent space risks losing crucial spatial details. Hence, inspired by \cite{moso}, all attention layers are applied in both spatial $(h \times w)$ and temporal $t$ dimensions. This ensures the model can capture not only the temporal relationships within the sequence but also the important spatial information embedded within each latent representation.

\subsubsection{Instance-level self-attention}
For each latent instance frame $z_k^t$ in the sequence, we first apply learnable positional embeddings. This embedding is added to the input features prior to self-attention to provide the model with information about the position of each instance within the sequence.
Scaled self-attention is then applied to each instance sequence separately in order to learn instance-specific dynamics:
\begin{equation}\label{eq:self_att}
\operatorname{SA}_{c}(\tilde{Z}_{k}) = \operatorname{softmax}\left(\frac{Q_{k}{K_k}^{T}}{\sqrt{d_{k}}}\right)V_k
\end{equation}
where $\operatorname{SA}$ denotes instance-specific self-attention for objects of class $c$, $Q_k, K_k^T$, which $T$ denotes transpose, and $V_k$ are the key, query and value calculated by a linear function on $\tilde{Z}_k$; $\frac{1}{\sqrt{d_k}}$ is a scaling factor that prevents excessively large values in the attention score. Following self-attention, we apply a further linear projection layer.

\subsubsection{Instance-level cross-attention}
After the self-attention layer that treats each instance separately, we apply cross-attention between instances to learn the potential relationships and interactions between objects. In this layer, each instance attend the space/time dimensions of each of the other instances:
\begin{equation}\label{eq:cross_att}
\operatorname{CA}(\tilde{Z}_k) = \bigoplus_{i=1\ldots N,\, i \neq k} \operatorname{softmax}\left(\frac{Q_{k}{K_i}^{T}}{\sqrt{d_{k}}}\right)V_i
\end{equation}
Here \(\operatorname{CA}\) denotes the cross-attention operation between instance \(k\) and the remaining instances. The value \(V_i\) and key \(K_i\) are derived from \(\tilde{Z}_i\), while the query originates from \(\tilde{Z}_k\). The cross-attention layer's output, being \(n-1\) times larger than the input because of concatenation, is reduced to the original size through a linear layer.

\begin{figure}[t]
  \centering
  \small 
  \setlength{\tabcolsep}{1pt} 
  \begin{tabular}{c cc|ccccc} 
    & \multicolumn{2}{c}{\centering{Input}} &   \multicolumn{5}{c}{\centering{Prediction}}   \\
    & \textbf{$t=1$} & \textbf{$t=5$} & \textbf{$t=7$} & \textbf{$t=9$} & \textbf{$t=13$} & \textbf{$t=18$} & \textbf{$t=30$}  \\
    \rotatebox{90}{GT} & \includegraphics[width=0.06\textwidth]{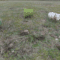} &
                                    \includegraphics[width=0.06\textwidth]{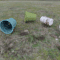} &
                                    \includegraphics[width=0.06\textwidth]{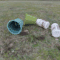 } &
                                    \includegraphics[width=0.06\textwidth]{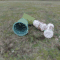 } &
                                    \includegraphics[width=0.06\textwidth]{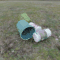} &
                                    \includegraphics[width=0.06\textwidth]{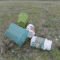} &
                                    \includegraphics[width=0.06\textwidth]{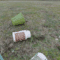} \\
     & \multicolumn{2}{c}{SiS}    &
                                 \includegraphics[width=0.06\textwidth]{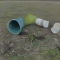 } &
                                 \includegraphics[width=0.06\textwidth]{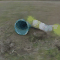 } &
                                 \includegraphics[width=0.06\textwidth]{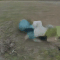} &
                                 \includegraphics[width=0.06\textwidth]{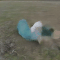} &
                                 \includegraphics[width=0.06\textwidth]{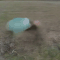} \\
     & \multicolumn{2}{c}{SNCAT}    &
                                 \includegraphics[width=0.06\textwidth]{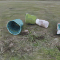 } &
                                 \includegraphics[width=0.06\textwidth]{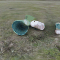 } &
                                 \includegraphics[width=0.06\textwidth]{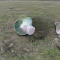} &
                                 \includegraphics[width=0.06\textwidth]{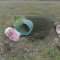} &
                                 \includegraphics[width=0.06\textwidth]{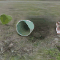} \\
    & \multicolumn{2}{c}{\textbf{SCAT}}   &
                                 \includegraphics[width=0.06\textwidth]{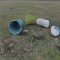 } &
                                 \includegraphics[width=0.06\textwidth]{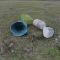 } &
                                 \includegraphics[width=0.06\textwidth]{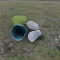} &
                                 \includegraphics[width=0.06\textwidth]{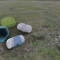} &
                                 \includegraphics[width=0.06\textwidth]{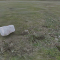} \\
  \end{tabular}
  \caption{Comparison of different model variants on the \textbf{Kubric-Real} dataset. SCAT successfully predicted that the blue pot bounced away whereas SNCAT neglected the interaction between other objects and let the blue pot go through from other objects. The single-slot model SiS fails to capture the appearances well, yielding indistinct predictions for later frames.}
  \label{ver_result}
  \vspace{-10pt}
\end{figure}

\subsubsection{Training and inference}
The model outputs probabilities over the codebook indices from OAAE, and we use cross-entropy loss to minimize the difference between the predicted and actual distributions.
During training, all model variants are trained with teacher forcing on 10-frame clips. Before the forward pass, 10\% noise sampled from a standard normal distribution \(\mathcal{N}(0, 1)\) is added to the input frames.
During inference, autoregressive sampling is used, starting from an initial sequence of conditioning frames, with softmax temperature treated as a hyperparameter.

\begin{table*}[t]
\centering \small
\caption{Quantitative results on \textbf{KTH} and \textbf{Real-Traffic} datasets}
\begin{tabular}{@{}lcccccccc@{}}
\toprule
                 & \multicolumn{4}{c}{\textbf{KTH}} & \multicolumn{4}{c}{\textbf{Real-Traffic}} \\ \cmidrule(r){2-5}  \cmidrule(l){6-9}
                 & PSNR$\uparrow$& SSIM$\uparrow$& LPIPS$\downarrow$& Num-Prms & PSNR$\uparrow$& SSIM$\uparrow$& LPIPS$\downarrow$ & Num-Prms \\
 \midrule
Single-Slot        & 25.26          &  0.765          &  0.110          & 48M &  28.34         & 0.933           & 0.026           & 286M \\
SNCAT              & 25.39          &   0.768         &  0.112          & 25M &  28.85         & 0.942           & 0.020           & 27M \\
SCAT               &\textbf{25.61}  & \textbf{0.774}  & \textbf{0.108}  & 23M &\textbf{30.04}  & \textbf{0.947}  & \textbf{0.017}  & 28M \\ 
\bottomrule
\end{tabular}
\label{tb:Weak_inter_quantitative}
    \vspace{-10pt}
\end{table*}

\begin{table*}[t]
    \centering \small
    \caption{Quantitative results on \textbf{CLEVR-2}, \textbf{CLEVR-3}, and \textbf{Kubric-Real} datasets}
    \setlength{\tabcolsep}{2pt}
    \begin{tabular}{@{}lcccccccccccc@{}}
    \toprule
    & \multicolumn{4}{c}{\textbf{CLEVR-2}} & \multicolumn{4}{c}{\textbf{CLEVR-3}} & \multicolumn{4}{c}{\textbf{Kubric-Real}} \\ 
    \cmidrule(r){2-5}  \cmidrule(lr){6-9} \cmidrule(l){10-13}
    & PSNR$\uparrow$ & SSIM$\uparrow$ & LPIPS$\downarrow$ &  Num-Prms & PSNR$\uparrow$ & SSIM$\uparrow$ & LPIPS$\downarrow$ &  Num-Prms & PSNR$\uparrow$ & SSIM$\uparrow$ & LPIPS$\downarrow$ & Num-Prms \\ 
    \midrule
    Single-Slot & \textbf{30.16} &   \textbf{0.906} &   0.061          & 105M &  30.15         &   0.895         & 0.068           & 186M &   22.88          & 0.729          &  0.165          &  287M  \\ 
    SNCAT       &   28.19        &   0.887          &   0.110          & 25M  &  28.01         &   0.879         & 0.104           & 26M  &   22.89          & 0.743          &  0.156          &  38M   \\ 
    SCAT        &  30.03         &   0.905          &  \textbf{0.056}  & 25M  & \textbf{32.25} & \textbf{0.925} & \textbf{0.030}  & 26M  &  \textbf{24.15}  & \textbf{0.773} & \textbf{0.122}  &  40M   \\
    \bottomrule
    \end{tabular}
    \label{tb:Strong_inter_quantitative}
    \vspace{-10pt}
\end{table*}

\subsubsection{Variants of the transformer}
We have described the transformer as used in the full model \textbf{SCAT}.
In the non-interacting model \textbf{SNCAT}, cross-attention is simply replaced by a per-object feed-forward network of similar capacity.
The single-slot version \textbf{SiS} has a single, larger latent vector for the whole scene instead of separate latents for each object, and we also increase the hidden dimensionality of the transformer (in fact resulting in considerably more parameters).
The number of feed-forward and self-attention layers remains the same.

\section{Experiments} \label{Experiments}

We perform a series of experiments to measure the benefit of separately modeling the dynamics of objects during video prediction.
Our focus is on comparing different model variants in a controlled setting, keeping model capacity approximately equal but changing whether the latent representation is decomposed over objects, and whether interactions between objects are modelled if so.
In addition, to place our results in context, we perform a comparative evaluation against other recent video prediction models under similar conditions.

\subsubsection{Experimental protocol}
Each model is given five frames as input, then predicts the following 5--25 frames depending on the dataset. We use $64\times64$ resolution for all datasets; further details on hyperparameters are in the appendix.
The models are implemented in PyTorch and trained on a single NVIDIA RTX 3090 GPU, reflecting our emphasis on computational efficiency and model scalability; further implementation details are given in the appendix. To ensure a rigorous comparison that focuses on the benefit of instance decomposition, we ensure the numbers of parameters in each model are as similar as possible.
Our focus is not on achieving state-of-the-art performance but rather on analyzing the benefits of explicit object-centric modeling within a balanced and controlled setting.
For quantitative results, we measure LPIPS \cite{LIPIPS}, PSNR \cite{PSNR} and SSIM \cite{SSIM}.
For brevity we report only average results in the main paper, and provide standard deviations in the appendix.

\subsubsection{Datasets}

We conduct experiments on five different datasets characterized by weak and strong interactions. We define weak interactions as scenarios where the dynamics of an instance are unaffected by other instances, or minimally so. In contrast, strong interactions involve instances significantly affecting each other's dynamics, such as during collisions.

The first weak interaction dataset we use is the \textbf{KTH} human action dataset \cite{KTH}. This includes six action types performed by 25 individuals. Although the primary focus is on the person, there remains some slight interaction between the person and the background, such as shadows cast by the individual on the background. Following MOSO \cite{moso}, we use videos of persons 1-16 for training and 17-25 for testing. We used \cite{lueddecke22clipseg} to segment the person and the background. Each model is given five frames and required to predict 15 future frames.
The second weak interaction dataset is the \textbf{Real-Traffic} dataset from \cite{relate2020}.
This comprises video clips taken from a CCTV camera overlooking a highway intersection. The background is static, and only the cars are moving in the scene; there are up to five cars per clip. The original dataset contains 615 video clips with various lengths, we split the dataset into a more standardized 10 frames per clip with 5089 clips for training and 2181 for validation. During inference the models are given five frames and required to predict five future frames. We used YOLOv8~\cite{yolov8} to extract each instance. 
Each car's motion is independent of other cars most of the time; however, interactions do occur, such as when a car stops before the intersection, causing other cars behind it to slow down. 
For quantitative evaluation, we therefore identify a subset of video clips from the test set with the strongest interactions. 
We calculate the distances between centroids of different cars, and select clips where the distance between any pair of cars is less than 25\% of the image size;
this yields a test set of 807 clips.

For strong interaction datasets, we used Kubric \cite{kubric} to generate a series of synthetic datasets inspired by CLEVRER \cite{clevrer} but exhibiting stronger interactions and more visual complexity. Full details on the dataset generation (and corresponding code) are included in the appendix.
Specifically, \textbf{CLEVR-2} contains scenes with two spheres with random velocity sampled such that they will collide; \textbf{CLEVR-3} scenes are similar but include another sphere that does not interact with the first two.
\textbf{Kubric-Real} uses a realistic background and replaces the basic geometric objects with 3D-scanned objects---bottles and pots since these exhibit interesting dynamics due to their cylindrical shapes.

\subsubsection{Internal and External Evaluation}

We first compare the different variations of our model, to evaluate the benefit of explicit object-centric modeling in a controlled setting.
Table \ref{tb:Weak_inter_quantitative} shows quantitative results on the two weak-interaction datasets. For \textbf{KTH}, the models are given five frames and required to predict 15 frames and for \textbf{Real-Traffic} they are required to predict five frames. In both datasets the proposed model performs best among other two variations. First, modeling the scene separately by segmenting it at the instance level (SNCAT) leads to predictions comparable to modeling the whole scene at once (Single-slot model), while using a much smaller model (25M vs. 48M parameters on \textbf{KTH}, 27M vs. 286M parameters on \textbf{Real-Traffic}). Second, adding cross-attention to the model to handle potential interactions between instances (SCAT) leads to an improvement in performance across all metrics. Since \textbf{KTH} features a single instance with negligible interaction, the performance improvement is subtle on each metric: SSIM (+0.006), PSNR (+0.35) and LPIPS (-0.02). On \textbf{Real-Traffic}, which has more instances and higher interactions, consistent improvements are observed in all metrics (PSNR: +1.70, SSIM: +0.014, LPIPS: -0.009). These results confirm the computational advantage of both the decomposition and cross-attention components of the approach.
\begin{figure*}[t]
  \centering
  \footnotesize 
  \setlength{\tabcolsep}{1pt} 
  \begin{tabular}{c cc|ccccc} 
    & \multicolumn{2}{c}{\centering{Input}} &   \multicolumn{5}{c}{\centering{Prediction}}   \\
    & \textbf{$t=1$} & \textbf{$5$} & \textbf{$7$} & \textbf{$10$} & \textbf{$13$} & \textbf{$16$} & \textbf{$20$}  \\
    \rotatebox{90}{\centering GT} & \includegraphics[width=0.035\textwidth]{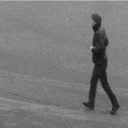} &
                                    \includegraphics[width=0.035\textwidth]{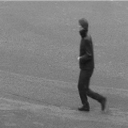} &
                                    \includegraphics[width=0.035\textwidth]{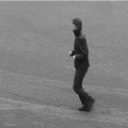} &
                                    \includegraphics[width=0.035\textwidth]{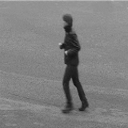} &
                                    \includegraphics[width=0.035\textwidth]{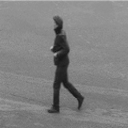} &
                                    \includegraphics[width=0.035\textwidth]{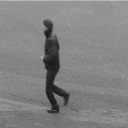} &
                                    \includegraphics[width=0.035\textwidth]{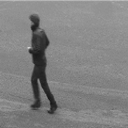} \\
     & \multicolumn{2}{c}{SVG}    &
                                 \includegraphics[width=0.035\textwidth]{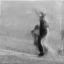 } &
                                 \includegraphics[width=0.035\textwidth]{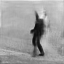 } &
                                 \includegraphics[width=0.035\textwidth]{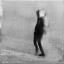} &
                                 \includegraphics[width=0.035\textwidth]{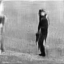} &
                                 \includegraphics[width=0.035\textwidth]{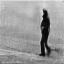} \\
    & \multicolumn{2}{c}{VideoGPT}   &
                                 \includegraphics[width=0.035\textwidth]{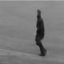 } &
                                 \includegraphics[width=0.035\textwidth]{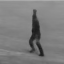 } &
                                 \includegraphics[width=0.035\textwidth]{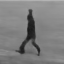} &
                                 \includegraphics[width=0.035\textwidth]{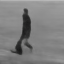} &
                                 \includegraphics[width=0.035\textwidth]{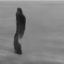} \\
    & \multicolumn{2}{c}{SimVP}   &
                                 \includegraphics[width=0.035\textwidth]{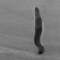 } &
                                 \includegraphics[width=0.035\textwidth]{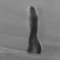 } &
                                 \includegraphics[width=0.035\textwidth]{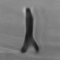} &
                                 \includegraphics[width=0.035\textwidth]{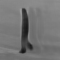} &
                                 \includegraphics[width=0.035\textwidth]{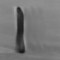} \\
    & \multicolumn{2}{c}{\textbf{Ours}}   &
                                 \includegraphics[width=0.035\textwidth]{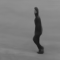 } &
                                 \includegraphics[width=0.035\textwidth]{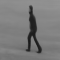 } &
                                 \includegraphics[width=0.035\textwidth]{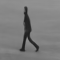} &
                                 \includegraphics[width=0.035\textwidth]{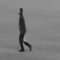} &
                                 \includegraphics[width=0.035\textwidth]{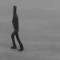} \\
  \end{tabular}
%
%
%
  \setlength{\tabcolsep}{1pt} 
  \begin{tabular}{cc|ccccc} 
    \multicolumn{2}{c}{\centering{Input}} &   \multicolumn{5}{c}{\centering{Prediction}}   \\
    \textbf{$1$} & \textbf{$5$} & \textbf{$6$} & \textbf{$7$} & \textbf{$8$} & \textbf{$9$} & \textbf{$10$}  \\
                         \includegraphics[width=0.035\textwidth]{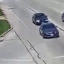} &
                         \includegraphics[width=0.035\textwidth]{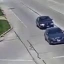} &
                         \includegraphics[width=0.035\textwidth]{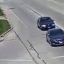} &
                         \includegraphics[width=0.035\textwidth]{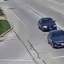} &
                         \includegraphics[width=0.035\textwidth]{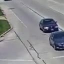} &
                         \includegraphics[width=0.035\textwidth]{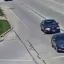} &
                         \includegraphics[width=0.035\textwidth]{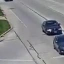} \\
    \multicolumn{2}{c}{}    &
                                 \includegraphics[width=0.035\textwidth]{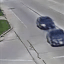} &
                                 \includegraphics[width=0.035\textwidth]{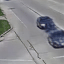} &
                                 \includegraphics[width=0.035\textwidth]{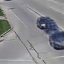} &
                                 \includegraphics[width=0.035\textwidth]{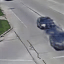} &
                                 \includegraphics[width=0.035\textwidth]{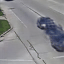} \\
    \multicolumn{2}{c}{}   &
                                 \includegraphics[width=0.035\textwidth]{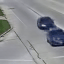} &
                                 \includegraphics[width=0.035\textwidth]{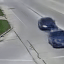} &
                                 \includegraphics[width=0.035\textwidth]{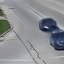} &
                                 \includegraphics[width=0.035\textwidth]{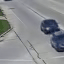} &
                                 \includegraphics[width=0.035\textwidth]{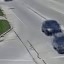} \\
    \multicolumn{2}{c}{}   &
                                 \includegraphics[width=0.035\textwidth]{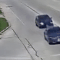} &
                                 \includegraphics[width=0.035\textwidth]{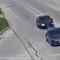} &
                                 \includegraphics[width=0.035\textwidth]{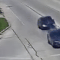} &
                                 \includegraphics[width=0.035\textwidth]{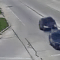} &
                                 \includegraphics[width=0.035\textwidth]{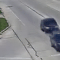} \\
    \multicolumn{2}{c}{}   &
                                 \includegraphics[width=0.035\textwidth]{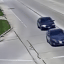} &
                                 \includegraphics[width=0.035\textwidth]{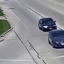} &
                                 \includegraphics[width=0.035\textwidth]{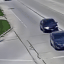} &
                                 \includegraphics[width=0.035\textwidth]{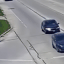} &
                                 \includegraphics[width=0.035\textwidth]{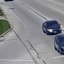} \\
  \end{tabular}
  \setlength{\tabcolsep}{1pt} 
  \begin{tabular}{cc|ccccc} 
    \multicolumn{2}{c}{\centering{Input}} &   \multicolumn{5}{c}{\centering{Prediction}}   \\
    \textbf{$1$} & \textbf{$5$} & \textbf{$7$} & \textbf{$9$} & \textbf{$13$} & \textbf{$18$} & \textbf{$30$}  \\
    \includegraphics[width=0.035\textwidth]{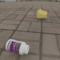} &
                                    \includegraphics[width=0.035\textwidth]{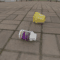} &
                                    \includegraphics[width=0.035\textwidth]{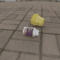 } &
                                    \includegraphics[width=0.035\textwidth]{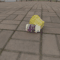 } &
                                    \includegraphics[width=0.035\textwidth]{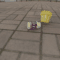} &
                                    \includegraphics[width=0.035\textwidth]{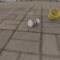} &
                                    \includegraphics[width=0.035\textwidth]{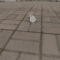} \\
    \multicolumn{2}{c}{}    &
                                 \includegraphics[width=0.035\textwidth]{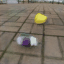 } &
                                 \includegraphics[width=0.035\textwidth]{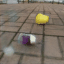 } &
                                 \includegraphics[width=0.035\textwidth]{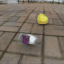} &
                                 \includegraphics[width=0.035\textwidth]{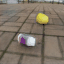} &
                                 \includegraphics[width=0.035\textwidth]{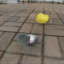} \\
    \multicolumn{2}{c}{}    &
                                 \includegraphics[width=0.035\textwidth]{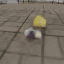 } &
                                 \includegraphics[width=0.035\textwidth]{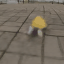 } &
                                 \includegraphics[width=0.035\textwidth]{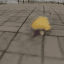} &
                                 \includegraphics[width=0.035\textwidth]{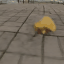} &
                                 \includegraphics[width=0.035\textwidth]{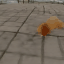} \\
    \multicolumn{2}{c}{}    &
                                 \includegraphics[width=0.035\textwidth]{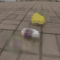 } &
                                 \includegraphics[width=0.035\textwidth]{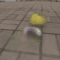 } &
                                 \includegraphics[width=0.035\textwidth]{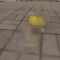} &
                                 \includegraphics[width=0.035\textwidth]{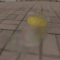} &
                                 \includegraphics[width=0.035\textwidth]{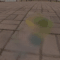} \\
    \multicolumn{2}{c}{}   &
                                 \includegraphics[width=0.035\textwidth]{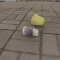 } &
                                 \includegraphics[width=0.035\textwidth]{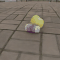 } &
                                 \includegraphics[width=0.035\textwidth]{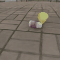} &
                                 \includegraphics[width=0.035\textwidth]{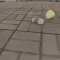} &
                                 \includegraphics[width=0.035\textwidth]{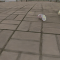} \\
  \end{tabular}
  \caption{Qualitative results from our full model and baselines on \textbf{KTH} (left), \textbf{Real-Traffic} (middle) and \textbf{Kubric-real} (right).}
  \label{fig:comparision with others}
    \vspace{-10pt}
\end{figure*}
\begin{table*}[t]
\centering \small
\setlength{\tabcolsep}{2pt}
\caption{Quantitative results on \textbf{KTH}, \textbf{Real-Traffic} and \textbf{Kubric-Real} datasets}
\begin{tabular}{@{}lcccccccccccc@{}}
\toprule
& \multicolumn{4}{c}{\textbf{KTH}} & \multicolumn{4}{c}{\textbf{Real-Traffic}}  & \multicolumn{4}{c}{\textbf{Kubric-Real}} \\
\cmidrule(r){2-5}\cmidrule(lr){6-9}\cmidrule(l){10-13}
 & PSNR$\uparrow$& SSIM$\uparrow$& LPIPS$\downarrow$& Num-Prms & PSNR$\uparrow$& SSIM$\uparrow$& LPIPS$\downarrow$ & Num-Prms & PSNR$\uparrow$ & SSIM$\uparrow$ & LPIPS$\downarrow$ & Num-Prms  \\ \midrule
SVG        &  16.27        & 0.572          &  0.160         & 23M &    23.67      &      0.893     &      0.101     & 31M & 15.48           & 0.615          &   0.720        & 41M \\ 
VideoGPT   &  24.39        & 0.788          & \textbf{0.086} & 41M &    29.19      &      0.928     &      0.023     & 55M & 23.68           & 0.702          &   0.153        & 67M \\ 
SimVP      &  25.45        & \textbf{0.810} &  0.117         & 56M &\textbf{30.51} &\textbf{0.949}  &      0.019     & 31M & 22.21           & 0.711          &   0.213        & 59M    \\ 
SCAT       &\textbf{25.52} & 0.774          &      0.108     & 23M &       30.04   &        0.947   & \textbf{0.017} & 28M & \textbf{24.15}  & \textbf{0.773} & \textbf{0.122} & 40M \\ \bottomrule
\end{tabular}

\label{tb:quantitative-others}
    \vspace{-10pt}
\end{table*}
Table~\ref{tb:Strong_inter_quantitative} provides quantitative results on the strong-interaction datasets.
On \textbf{CLEVR-2}, the SCAT model (PSNR: 30.03) performs similarly to the single-slot model (PSNR: 30.16) but outperforms it on LPIPS (0.056 vs. 0.061). In contrast, SNCAT performs worse than the single-slot model both on \textbf{CLEVR-2} and \textbf{CLEVR-3} datasets, this due to the lack of cross-attention to model interactions between objects.
In \textbf{Kubric-Real}, SNCAT preserves object shapes better than the single-slot model, which struggles with deformation after collision. SCAT outperforms both models in LPIPS (0.122 vs. 0.165 for the single-slot model) and SSIM (0.773 vs. 0.729 for the single-slot model), emphasizing the importance of cross-attention in more realistic and complex interaction scenes.
These results confirm our hypothesis that instance segmentation is important for video prediction and that cross-attention is an effective way to encode strong interactions. Moreover, without cross-attention, instance separation on its own is sufficient to achieve similar or better performance compared to the baseline single-slot model on complex scenes (\textbf{Real-traffic}, \textbf{Kubric-Real}) having more than two instances, with only a fraction of the parameters.  
Although the main focus of our work is on measuring the benefit of object-centric video modeling in a controlled setting, we also compare our method with other similar state-of-the-art methods to better contextualize those results. Our model is designed to be small yet efficient, demonstrating high performance without the need for large-scale resources.
In contrast, many existing models rely on significantly larger architectures to achieve similar results, which can be resource-intensive and less practical.
To ensure a fair and balanced evaluation, we therefore adjusted each method's hyperparameters to match our model's size (i.e.~number of weights), providing a level playing field for comparison.
We compare against VideoGPT \cite{yan2021videogpt}, which uses a similar architecture, and the CNN-based SimVP \cite{SimVP} for a comprehensive evaluation.

Prediction performance on \textbf{KTH}, \textbf{Real-Traffic} and \textbf{Kubric-Real} are presented in Table~\ref{tb:quantitative-others}. 
The SCAT model outperforms or is competitive with other models across all three datasets, even with its smaller model size, confirming the effectiveness of instance-level segmentation and cross-attention.
On the simpler \textbf{KTH} dataset, VideoGPT achieves slightly higher SSIM than SCAT (0.788 vs 0.772) and slightly lower LPIPS than SCAT (0.086 vs 0.098), but lower quality according to PSNR (24.39 vs 25.46). Moreover, from Fig.\ref{fig:comparision with others} we can see that only SCAT maintained human posture throughout the prediction.
On \textbf{Real-Traffic}, SCAT again achieves competitive performance, with PSNR of 30.04, which is higher than VideoGPT (29.19) but lower than SimVP (30.51). SCAT performs best under the perceptually robust LPIPS metric (0.017), outperforming both VideoGPT (0.023) and SimVP (0.019), indicating better perceptual quality. Also, from Figure~\ref{fig:comparision with others} we can see that when $t$=9 and $t$=10, SCAT maintained the distance between two cars and kept them separate while the other models merged two cars. 
On \textbf{Kubric-Real}, where strong interactions and realistic objects are present, our model leads by a large margin on every metric.
This further shows the proposed model's improvement is larger on scenes with more instances and strong interactions. In Figure \ref{fig:comparision with others} other methods failed to predict the collision between two objects, while SCAT predicted this accurately and maintained the object shape.

\section{Conclusion}

In this paper, we investigated and analyzed the benefits of explicit object-centric decomposition in video prediction. We proposed a video prediction pipeline based on an object-aware VQ-VAE and multi-object Transformer, that operates on separate objects extracted via panoptic segmentation; we also defined variants that lack object-decomposition and support for interactions. We evaluated the proposed models on five datasets, finding that when a dynamic scene is explicitly decomposed and encoded into a structured latent vector, prediction quality is better than an equal-capacity model without decomposition. 

\bibliography{main}


\appendix
\section*{Implementation Details}
Specific implementation details of both \textbf{Object Aware Auto-Encoder (OAAE)}, \textbf{Stochastic Class-Attended Transformer (SCAT)} and their variants are given in table \ref{tab:ooae-ipml-detrail} and \ref{tab:scat-ipml-detrail}. 
We ensured that the non-decomposed version was fairly compared to the decomposed version by adjusting the embedding dimensions accordingly.
Specifically, the embedding dimension in the non-decomposed version was set to be $N$ times larger than the embedding dimension of a single instance in the decomposed setting, where $N$ represents the total number of instances.
For example, in the \textbf{Kubric-Real} dataset, there are three classes: background, bottles, and pots. The background class is assigned one slot, the bottles class is assigned two slots, and the pots class is assigned two slots, totaling five instances. 
Thus, if each instance in the decomposed version has an embedding dimension of 128, then in the non-decomposed version, the embedding dimension is set to 640, which is 128 multiplied by the total number of instances (5).
\begin{table*}[t]
\centering \small
\setlength{\tabcolsep}{2pt}
\begin{tabular}{@{}lccccccccccc@{}}
\toprule
  & \multicolumn{2}{c}{\textbf{KTH}} & \multicolumn{2}{c}{\textbf{Real-Traffic}} & \multicolumn{2}{c}{\textbf{CLEVR-2}} & \multicolumn{2}{c}{\textbf{CLEVR-3}} & \multicolumn{2}{c}{\textbf{Kubric-Real}} \\
 \cmidrule(r){2-3}\cmidrule(lr){4-5}\cmidrule(l){6-7}\cmidrule(l){8-9}\cmidrule(l){10-11}
    & OAAE & Non-Decom & OAAE & Non-Decom & OAAE & Non-Decom & OAAE & Non-Decom & OAAE & Non-Decom \\
 \midrule
In Channels                    & \multicolumn{2}{c}{1} & \multicolumn{2}{c}{3} & \multicolumn{2}{c}{3} & \multicolumn{2}{c}{3} & \multicolumn{2}{c}{3} \\
Num Instance                   & 2 & 1 & 5 & 1 & 3 & 1 & 4 & 1 & 5 & 1 \\ 
Num Classes                    & 2 & 1 & 2 & 1 & 2 & 1 & 2 & 1 & 3 & 1\\
Embedding Dim Per Instance     & 128 & 256 & 128 & 640 & 128 & 384 & 128 & 512 & 128 & 640 \\
Num Embeddings                 & \multicolumn{2}{c}{5120} & \multicolumn{2}{c}{5120} & \multicolumn{2}{c}{5120} & \multicolumn{2}{c}{5120} & \multicolumn{2}{c}{5120} &\\
Conv Hidden Dims               & \multicolumn{2}{c}{128, 256} & \multicolumn{2}{c}{128, 256} & \multicolumn{2}{c}{128, 256} & \multicolumn{2}{c}{128, 256} & \multicolumn{2}{c}{128, 256} \\
Num Residual Layers            & \multicolumn{2}{c}{6} & \multicolumn{2}{c}{6} & \multicolumn{2}{c}{6} & \multicolumn{2}{c}{6} & \multicolumn{2}{c}{6} \\
Batch Size                     & \multicolumn{2}{c}{8} & \multicolumn{2}{c}{8} & \multicolumn{2}{c}{8} & \multicolumn{2}{c}{8} & \multicolumn{2}{c}{8} &\\
Learning Rate                  & \multicolumn{2}{c}{$10^{-4}$} & \multicolumn{2}{c}{$10^{-4}$} & \multicolumn{2}{c}{$10^{-4}$} & \multicolumn{2}{c}{$10^{-4}$} & \multicolumn{2}{c}{$10^{-4}$} & \\
\bottomrule
\end{tabular}
    \vspace{-4pt}
\caption{HyperParameters of OAAE on \textbf{KTH}, \textbf{Real-Traffic}, \textbf{CLEVR-2}, \textbf{CLEVR-3}  and \textbf{Kubric-Real} Datasets}
\label{tab:ooae-ipml-detrail}
\end{table*}
\begin{table*}[t]
\centering \small
\setlength{\tabcolsep}{2pt}
\begin{tabular}{@{}lcccccccccccccccc@{}}
\toprule
  & \multicolumn{3}{c}{\textbf{KTH}} & \multicolumn{3}{c}{\textbf{Real-Traffic}} & \multicolumn{3}{c}{\textbf{CLEVR-2}} & \multicolumn{3}{c}{\textbf{CLEVR-3}} & \multicolumn{3}{c}{\textbf{Kubric-Real}} \\
 \cmidrule(r){2-4}\cmidrule(lr){5-7}\cmidrule(l){8-10}\cmidrule(l){11-13}\cmidrule(l){14-16}
    & SCAT & SNCAT & SiS & SCAT & SNCAT & SiS & SCAT & SNCAT & SiS & SCAT & SNCAT & SiS & SCAT & SNCAT & SiS \\
 \midrule
Num Instance                   & \multicolumn{2}{c}{2} & 1 & \multicolumn{2}{c}{5} & 1 & \multicolumn{2}{c}{3} & 1 & \multicolumn{2}{c}{4} & 1 & \multicolumn{2}{c}{5} & 1 \\ 
Num Classes                    & \multicolumn{2}{c}{2} & 1 & \multicolumn{2}{c}{2} & 1 & \multicolumn{2}{c}{2} & 1 & \multicolumn{2}{c}{2} & 1 & \multicolumn{2}{c}{3} & 1\\
VQVAE Dim                      & \multicolumn{2}{c}{128} & 256 & \multicolumn{2}{c}{128} & 640 & \multicolumn{2}{c}{128} & 384 & \multicolumn{2}{c}{128} & 512 & \multicolumn{2}{c}{128} & 640 \\
Embedding Dim Per Instance     & \multicolumn{2}{c}{256} & 512 & \multicolumn{2}{c}{256} & 1280 & \multicolumn{2}{c}{256} & 768 & \multicolumn{2}{c}{256} & 1024 & \multicolumn{2}{c}{256} & 1280 \\
Num Attention Head             & \multicolumn{3}{c}{16} & \multicolumn{3}{c}{16} & \multicolumn{3}{c}{16} & \multicolumn{3}{c}{16} & \multicolumn{3}{c}{16} & \\
FeedForward expanding Factor   & \multicolumn{3}{c}{2} &  \multicolumn{3}{c}{2} &  \multicolumn{3}{c}{2} &  \multicolumn{3}{c}{2} &  \multicolumn{3}{c}{2} & \\
Depth                          & \multicolumn{3}{c}{4} & \multicolumn{3}{c}{4} & \multicolumn{3}{c}{4} & \multicolumn{3}{c}{4} & \multicolumn{3}{c}{4} & \\
Drop Out                       & \multicolumn{3}{c}{0.3} & \multicolumn{3}{c}{0.3} & \multicolumn{3}{c}{0.3} & \multicolumn{3}{c}{0.3} & \multicolumn{3}{c}{0.3} & \\
Batch Size                     & \multicolumn{3}{c}{1} & \multicolumn{3}{c}{1} & \multicolumn{3}{c}{1} & \multicolumn{3}{c}{1} & \multicolumn{3}{c}{1} & \\
Learning Rate                  & \multicolumn{3}{c}{$10^{-4}$} & \multicolumn{3}{c}{$10^{-4}$} & \multicolumn{3}{c}{$10^{-4}$} & \multicolumn{3}{c}{$10^{-4}$} & \multicolumn{3}{c}{$10^{-4}$} & \\
LR Scheduler                   & \multicolumn{3}{c}{Cosine} & \multicolumn{3}{c}{Cosine} & \multicolumn{3}{c}{Cosine} & \multicolumn{3}{c}{Cosine} & \multicolumn{3}{c}{Cosine} & \\
Warm-up Steps                  & \multicolumn{3}{c}{10000} & \multicolumn{3}{c}{10000} & \multicolumn{3}{c}{10000} & \multicolumn{3}{c}{10000} & \multicolumn{3}{c}{10000} & \\

 \bottomrule
\end{tabular}
    \vspace{-4pt}
\caption{HyperParameters of SCAT and its variants on \textbf{KTH}, \textbf{Real-Traffic}, \textbf{CLEVR-2}, \textbf{CLEVR-3}  and \textbf{Kubric-Real} Datasets}
\label{tab:scat-ipml-detrail}
\end{table*}
\section*{Dataset Details}
\subsubsection{Decomposition}
For KTH, we use CLIPSeg with the prompt \texttt{'person'} and \texttt{'background'} to decompose the frames. For Real-Traffic, we use YOLOv8 to be our instance segmentor. For Kubric generated datasets, because the instance segmentation map is available with the generation, we directly use these to extract the instances.
\subsubsection{Synthetic Datasets Generation}
We use Kubric to generate \textbf{CLEVR-2}, \textbf{CLEVR-3} and \textbf{Kubric-Real}. Genration parameters are given in Table \ref{tab:gen-detail}. The table outlines the parameters for generating the CLEVR-2, CLEVR-3, and Kubric-Real datasets. All three datasets use a colliding position range of \([-1, 1]\) and a fixed, static camera looking at \((0, 0)\). The summoning radius is set to 5 for CLEVR datasets and 8 for Kubric-Real, with minimum summoning distances of 2 for CLEVR and 4 for Kubric-Real. CLEVR datasets feature object friction values of 0.4 for metal spheres and 0.8 for rubber spheres, while Kubric-Real has a uniform friction of 1.0. This higher friction in Kubric-Real necessitates a larger maximum initial velocity of 7, compared to 5 in the CLEVR datasets. The number of objects also increases from 2 in CLEVR-2 to 3 in CLEVR-3, and 4 in Kubric-Real. More details are given in table \ref{tab:gen-detail}.
\section*{More Results}
Here we illustrate the quantitative results of our models on each datasets. The results are produced by sampling using set of 10 temperature values from 0.1 to 0.9 meaning that lower to higher stochasticity, then the best result is selected to compare. Each dataset's testing set is divided into 10 subsets. First, the mean of each metric on the subset is calculated, then the overall mean and std is calculated using 10 mean values accordingly.
\begin{figure}
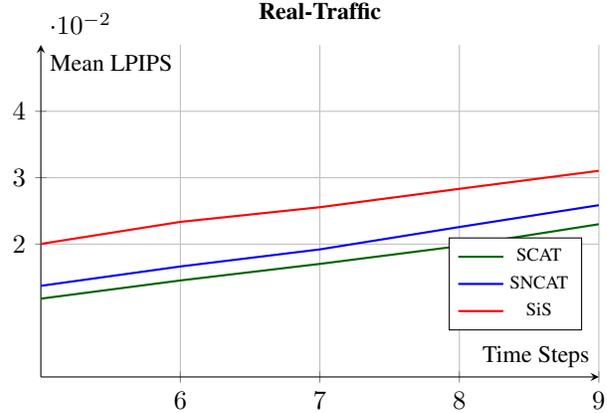


    \caption{Comparison of Mean LPIPS Values on \textbf{Real-Traffic} datasets}
    \label{fig:mean-rf}
\end{figure}
\begin{table}[t]
\centering \small
\setlength{\tabcolsep}{2pt}
\begin{tabular}{@{}lcccc@{}}
\toprule
    &  \textbf{CLEVR-2} & \textbf{CLEVR-3} & \textbf{Kubric-Real} \\
    \midrule
    Colliding Position Range $(x, y)$ & [-1,1]        & [-1,1]       & [-1,1]       \\ 
    Radius for Summoning Objects      &   5           &   5          & 8            \\
    Min Distance When Summoning       &   2           &   2          & 4            \\
    Max Initial Velocity              &   5           &   5          & 7            \\
    Ground Friction                   &   0.3         &   0.3        & 0.3          \\
    Object Friction                   & 0.4,0.8       & 0.4,0.8      & 1.0          \\
    Num Objects                       & 2             & 3            & 4            \\
    Num Object Class                  & 1             & 1            & 2            \\
    Camera Position                   & Fixed Static  & Fixed Static & Fixed Static \\
    Camera Looks At $(x, y, z)$       & (0, 0, 0)     & (0, 0, 0)    & (0, 0, 0)    \\
    \bottomrule
\end{tabular}
    \vspace{-4pt}
\caption{Parameters for Generating \textbf{CLEVR-2}, \textbf{CLEVR-3} and \textbf{Kubric-Real} Datasets}
\label{tab:gen-detail}
\end{table}
%
%
\begin{figure}[t]
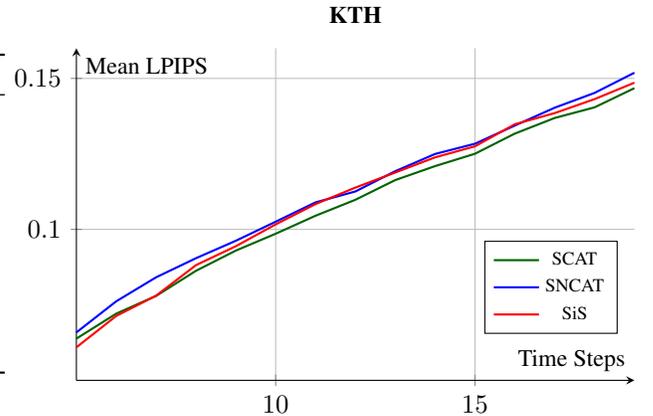

    \centering
    %
    \caption{Comparison of Mean LPIPS Values on \textbf{KTH} datasets}
    
\end{figure}
\begin{figure*}[t]
    \centering
    %
    \caption{Comparison of Mean LPIPS Values on \textbf{CLEVR-2} dataset}
    \label{fig:clevr-2-combined-mean}
\end{figure*}
\begin{figure*}[t]
    \centering
    %
    \caption{Comparison of Mean LPIPS Values on \textbf{CLEVR-3} dataset}
\end{figure*}
\begin{figure*}[t]
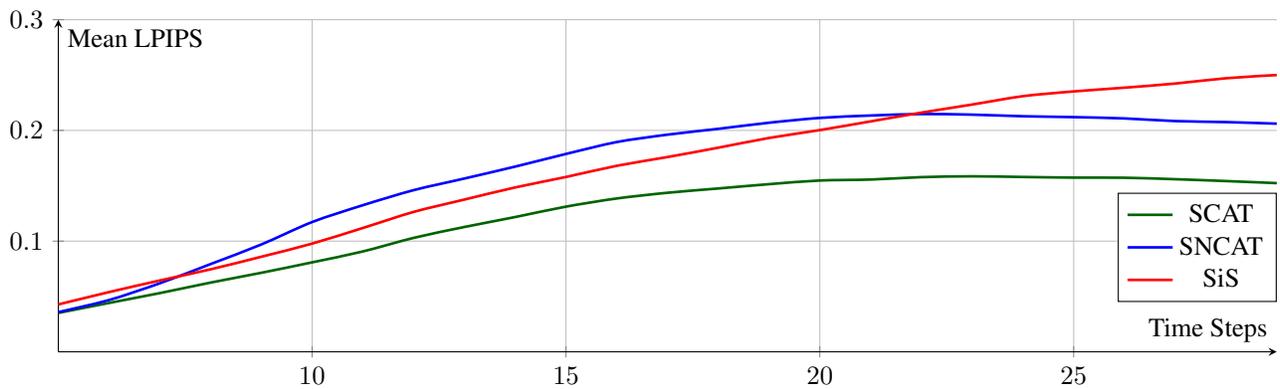

    \centering
    %
    \caption{Comparison of Mean LPIPS Values on \textbf{Kubric-Real} dataset}
    \label{fig:kubric-real-mean}
\end{figure*}
On \textbf{Kubric-Real} dataset, by reaching the end of prediction \textbf{SCAT} and \textbf{SNCAT} starts to predict better. This is because in the ground truth, the objects are either no longer in the scene or staying still. 
%
\begin{figure*}[t]
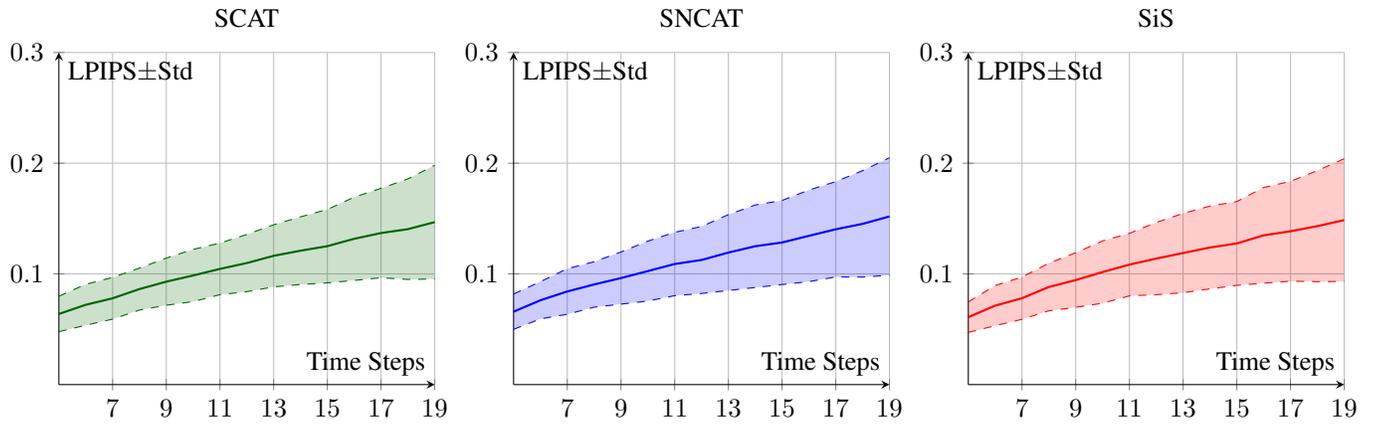

    \centering
    %
    \caption{Mean and Std (Standard Deviation) of LPIPS metric on \textbf{KTH} dataset}
\end{figure*}
\begin{figure*}[t]
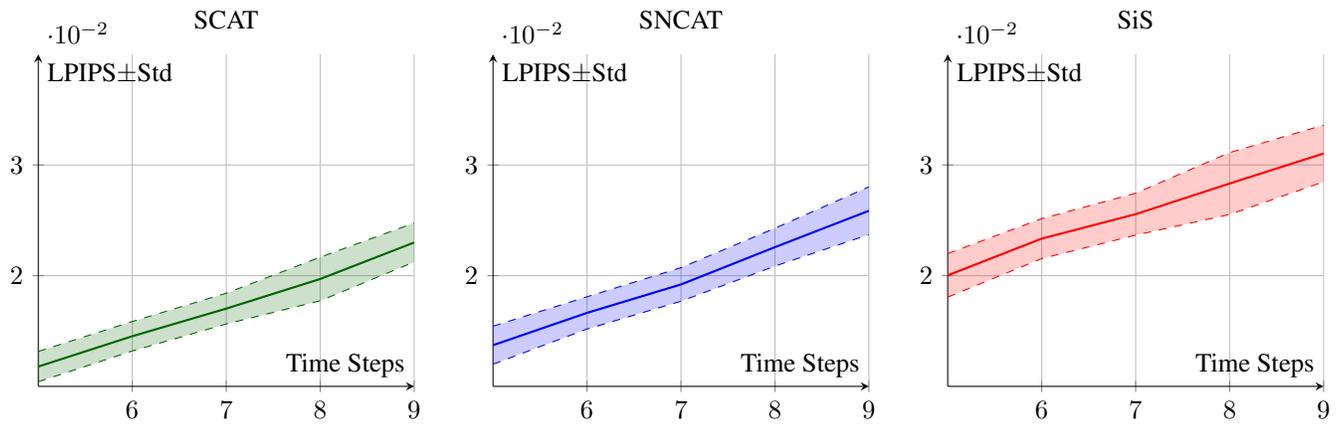

    \centering
    %
    \caption{Mean and Std (Standard Deviation) of LPIPS metric on \textbf{Real-Traffic} dataset}
    \label{fig:rf}
\end{figure*}
\begin{figure*}[t]
    \centering
    %
    \caption{Mean and Std of LPIPS metric on \textbf{CLEVR-2} dataset}
    \label{fig:clevr-2}
\end{figure*}
\begin{figure*}[t]
    \centering
    %
    \caption{Mean and Std of LPIPS metric on \textbf{CLEVR-3} dataset}
    \label{fig:clevr-3}
\end{figure*}
\begin{figure*}[t]
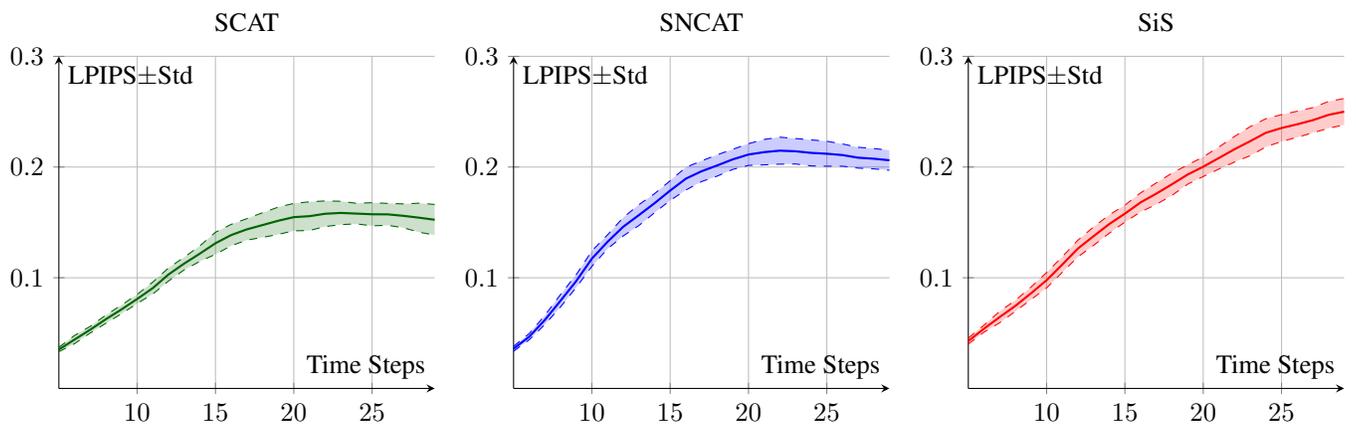

    \centering
    %
    \caption{Mean and Std of LPIPS metric on \textbf{Kubric-Real} dataset}
    \label{fig:kubric-real}
\end{figure*}
\begin{figure*}[t]
  \centering
  \small 
  \setlength{\tabcolsep}{1pt} 
  \begin{tabular}{c cc|ccccccc} 
    & \multicolumn{2}{c}{\centering{Input}} &   \multicolumn{7}{c}{\centering{Prediction}}   \\
    & \textbf{$t=1$} & \textbf{$t=5$} & \textbf{$t=7$} & \textbf{$t=9$} & \textbf{$t=11$} & \textbf{$t=13$} &\textbf{$t=15$} & \textbf{$t=18$} & \textbf{$t=25$}  \\
    \rotatebox{90}{GT} & \includegraphics[width=0.1\textwidth]{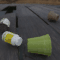} &
                                    \includegraphics[width=0.1\textwidth]{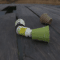} &
                                    \includegraphics[width=0.1\textwidth]{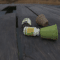 } &
                                    \includegraphics[width=0.1\textwidth]{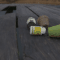 } &
                                    \includegraphics[width=0.1\textwidth]{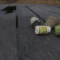 } &
                                    \includegraphics[width=0.1\textwidth]{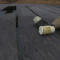} &
                                    \includegraphics[width=0.1\textwidth]{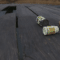} &
                                    \includegraphics[width=0.1\textwidth]{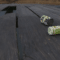} &
                                    \includegraphics[width=0.1\textwidth]{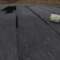} \\
\end{tabular}
  \caption{\textbf{Kubric-Real} Example 1}
  \label{Kubric-exmaple-1}
\end{figure*}
\begin{figure*}[t]
  \centering
  \small 
  \setlength{\tabcolsep}{1pt} 
  \begin{tabular}{c cc|ccccccc} 
    & \multicolumn{2}{c}{\centering{Input}} &   \multicolumn{7}{c}{\centering{Prediction}}   \\
    & \textbf{$t=1$} & \textbf{$t=5$} & \textbf{$t=7$} & \textbf{$t=9$} & \textbf{$t=11$} & \textbf{$t=13$} &\textbf{$t=15$} & \textbf{$t=18$} & \textbf{$t=25$}  \\
    \rotatebox{90}{GT}            & \includegraphics[width=0.1\textwidth]{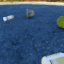} &
                                    \includegraphics[width=0.1\textwidth]{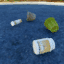} &
                                    \includegraphics[width=0.1\textwidth]{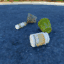 } &
                                    \includegraphics[width=0.1\textwidth]{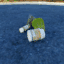 } &
                                    \includegraphics[width=0.1\textwidth]{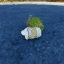 } &
                                    \includegraphics[width=0.1\textwidth]{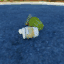} &
                                    \includegraphics[width=0.1\textwidth]{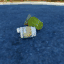} &
                                    \includegraphics[width=0.1\textwidth]{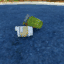} &
                                    \includegraphics[width=0.1\textwidth]{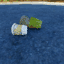} \\
\end{tabular}
  \caption{\textbf{Kubric-Real} Example 2}
  \label{Kubric-exmaple-2}
\end{figure*}
\begin{figure*}[t]
  \centering
  \small 
  \setlength{\tabcolsep}{1pt} 
  \begin{tabular}{c cc|ccccccc} 
    & \multicolumn{2}{c}{\centering{Input}} &   \multicolumn{7}{c}{\centering{Prediction}}   \\
    & \textbf{$t=1$} & \textbf{$t=5$} & \textbf{$t=7$} & \textbf{$t=9$} & \textbf{$t=11$} & \textbf{$t=13$} &\textbf{$t=15$} & \textbf{$t=18$} & \textbf{$t=25$}  \\
    \rotatebox{90}{GT}            & \includegraphics[width=0.1\textwidth]{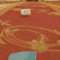} &
                                    \includegraphics[width=0.1\textwidth]{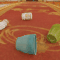} &
                                    \includegraphics[width=0.1\textwidth]{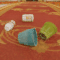 } &
                                    \includegraphics[width=0.1\textwidth]{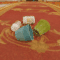 } &
                                    \includegraphics[width=0.1\textwidth]{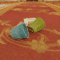 } &
                                    \includegraphics[width=0.1\textwidth]{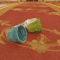} &
                                    \includegraphics[width=0.1\textwidth]{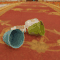} &
                                    \includegraphics[width=0.1\textwidth]{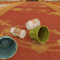} &
                                    \includegraphics[width=0.1\textwidth]{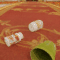} \\
\end{tabular}
  \caption{\textbf{Kubric-Real} Example 3}
  \label{Kubric-exmaple-3}
\end{figure*}
\begin{figure*}[t]
  \centering
  \small 
  \setlength{\tabcolsep}{1pt} 
  \begin{tabular}{c cc|ccccccc} 
    & \multicolumn{2}{c}{\centering{Input}} &   \multicolumn{7}{c}{\centering{Prediction}}   \\
    & \textbf{$t=1$} & \textbf{$t=5$} & \textbf{$t=7$} & \textbf{$t=9$} & \textbf{$t=11$} & \textbf{$t=13$} &\textbf{$t=15$} & \textbf{$t=18$} & \textbf{$t=25$}  \\
    \rotatebox{90}{GT}            & \includegraphics[width=0.1\textwidth]{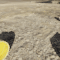} &
                                    \includegraphics[width=0.1\textwidth]{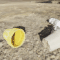} &
                                    \includegraphics[width=0.1\textwidth]{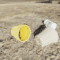 } &
                                    \includegraphics[width=0.1\textwidth]{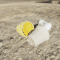 } &
                                    \includegraphics[width=0.1\textwidth]{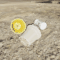 } &
                                    \includegraphics[width=0.1\textwidth]{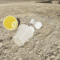} &
                                    \includegraphics[width=0.1\textwidth]{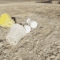} &
                                    \includegraphics[width=0.1\textwidth]{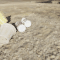} &
                                    \includegraphics[width=0.1\textwidth]{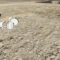} \\
\end{tabular}
  \caption{\textbf{Kubric-Real} Example 4}
  \label{Kubric-exmaple-4}
\end{figure*}
\begin{figure*}[t]
  \centering
  \small 
  \setlength{\tabcolsep}{1pt} 
  \begin{tabular}{c cc|ccccccc} 
    & \multicolumn{2}{c}{\centering{Input}} &   \multicolumn{7}{c}{\centering{Prediction}}   \\
    & \textbf{$t=1$} & \textbf{$t=5$} & \textbf{$t=7$} & \textbf{$t=9$} & \textbf{$t=11$} & \textbf{$t=13$} &\textbf{$t=15$} & \textbf{$t=18$} & \textbf{$t=25$}  \\
    \rotatebox{90}{GT}            & \includegraphics[width=0.1\textwidth]{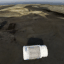} &
                                    \includegraphics[width=0.1\textwidth]{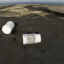} &
                                    \includegraphics[width=0.1\textwidth]{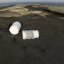 } &
                                    \includegraphics[width=0.1\textwidth]{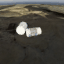 } &
                                    \includegraphics[width=0.1\textwidth]{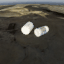 } &
                                    \includegraphics[width=0.1\textwidth]{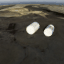} &
                                    \includegraphics[width=0.1\textwidth]{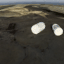} &
                                    \includegraphics[width=0.1\textwidth]{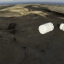} &
                                    \includegraphics[width=0.1\textwidth]{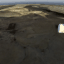} \\
\end{tabular}
  \caption{\textbf{Kubric-Real} Example 5}
  \label{Kubric-exmaple-5}
\end{figure*}
\begin{figure*}[t]
  \centering
  \small 
  \setlength{\tabcolsep}{1pt} 
  \begin{tabular}{c cc|ccccccc} 
    & \multicolumn{2}{c}{\centering{Input}} &   \multicolumn{5}{c}{\centering{Prediction}}   \\
    & \textbf{$t=1$} & \textbf{$t=5$} & \textbf{$t=6$} & \textbf{$t=7$} & \textbf{$t=8$} & \textbf{$t=9$} &\textbf{$t=10$}  \\
    \rotatebox{90}{GT}            & \includegraphics[width=0.12\textwidth]{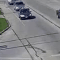} &
                                    \includegraphics[width=0.12\textwidth]{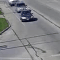} &
                                    \includegraphics[width=0.12\textwidth]{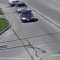 } &
                                    \includegraphics[width=0.12\textwidth]{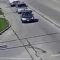 } &
                                    \includegraphics[width=0.12\textwidth]{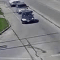 } &
                                    \includegraphics[width=0.12\textwidth]{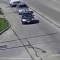} &
                                    \includegraphics[width=0.12\textwidth]{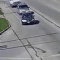} \\
     & \multicolumn{2}{c}{SiS}    &
                                 \includegraphics[width=0.12\textwidth]{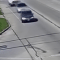} &
                                 \includegraphics[width=0.12\textwidth]{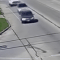} &
                                 \includegraphics[width=0.12\textwidth]{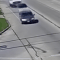 } &
                                 \includegraphics[width=0.12\textwidth]{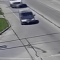} &
                                 \includegraphics[width=0.12\textwidth]{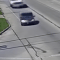} \\
\end{tabular}
  \caption{\textbf{Real-Traffic} Example 1}
  \label{rf-exmaple-1}
\end{figure*}
\begin{figure*}[t]
  \centering
  \small 
  \setlength{\tabcolsep}{1pt} 
  \begin{tabular}{c cc|ccccccc} 
    & \multicolumn{2}{c}{\centering{Input}} &   \multicolumn{5}{c}{\centering{Prediction}}   \\
    & \textbf{$t=1$} & \textbf{$t=5$} & \textbf{$t=6$} & \textbf{$t=7$} & \textbf{$t=8$} & \textbf{$t=9$} &\textbf{$t=10$}  \\
    \rotatebox{90}{GT}            & \includegraphics[width=0.12\textwidth]{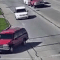} &
                                    \includegraphics[width=0.12\textwidth]{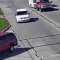} &
                                    \includegraphics[width=0.12\textwidth]{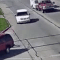 } &
                                    \includegraphics[width=0.12\textwidth]{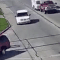 } &
                                    \includegraphics[width=0.12\textwidth]{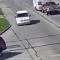 } &
                                    \includegraphics[width=0.12\textwidth]{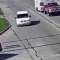} &
                                    \includegraphics[width=0.12\textwidth]{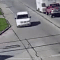} \\
     & \multicolumn{2}{c}{SiS}    &
                                 \includegraphics[width=0.12\textwidth]{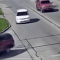} &
                                 \includegraphics[width=0.12\textwidth]{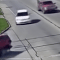} &
                                 \includegraphics[width=0.12\textwidth]{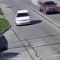 } &
                                 \includegraphics[width=0.12\textwidth]{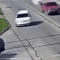} &
                                 \includegraphics[width=0.12\textwidth]{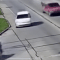} \\
\end{tabular}
  \caption{\textbf{Real-Traffic} Example 2}
  \label{rf-exmaple-2}
\end{figure*}
\begin{figure*}[t]
  \centering
  \small 
  \setlength{\tabcolsep}{1pt} 
  \begin{tabular}{c cc|ccccccc} 
    & \multicolumn{2}{c}{\centering{Input}} &   \multicolumn{5}{c}{\centering{Prediction}}   \\
    & \textbf{$t=1$} & \textbf{$t=5$} & \textbf{$t=6$} & \textbf{$t=7$} & \textbf{$t=8$} & \textbf{$t=9$} &\textbf{$t=10$}  \\
    \rotatebox{90}{GT}            & \includegraphics[width=0.12\textwidth]{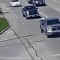} &
                                    \includegraphics[width=0.12\textwidth]{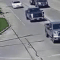} &
                                    \includegraphics[width=0.12\textwidth]{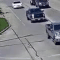 } &
                                    \includegraphics[width=0.12\textwidth]{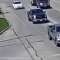 } &
                                    \includegraphics[width=0.12\textwidth]{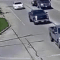 } &
                                    \includegraphics[width=0.12\textwidth]{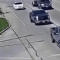} &
                                    \includegraphics[width=0.12\textwidth]{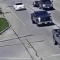} \\
     & \multicolumn{2}{c}{SiS}    &
                                 \includegraphics[width=0.12\textwidth]{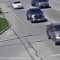} &
                                 \includegraphics[width=0.12\textwidth]{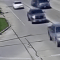} &
                                 \includegraphics[width=0.12\textwidth]{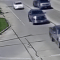 } &
                                 \includegraphics[width=0.12\textwidth]{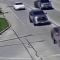} &
                                 \includegraphics[width=0.12\textwidth]{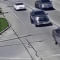} \\
\end{tabular}
  \caption{\textbf{Real-Traffic} Example 3}
  \label{rf-exmaple-3}
\end{figure*}
\begin{figure*}[t]
  \centering
  \small 
  \setlength{\tabcolsep}{1pt} 
  \begin{tabular}{c cc|ccccccc} 
    & \multicolumn{2}{c}{\centering{Input}} &   \multicolumn{5}{c}{\centering{Prediction}}   \\
    & \textbf{$t=1$} & \textbf{$t=5$} & \textbf{$t=6$} & \textbf{$t=7$} & \textbf{$t=8$} & \textbf{$t=9$} &\textbf{$t=10$}  \\
    \rotatebox{90}{GT}            & \includegraphics[width=0.12\textwidth]{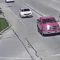} &
                                    \includegraphics[width=0.12\textwidth]{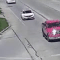} &
                                    \includegraphics[width=0.12\textwidth]{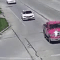 } &
                                    \includegraphics[width=0.12\textwidth]{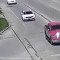 } &
                                    \includegraphics[width=0.12\textwidth]{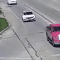 } &
                                    \includegraphics[width=0.12\textwidth]{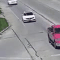} &
                                    \includegraphics[width=0.12\textwidth]{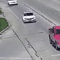} \\
     & \multicolumn{2}{c}{SiS}    &
                                 \includegraphics[width=0.12\textwidth]{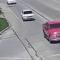} &
                                 \includegraphics[width=0.12\textwidth]{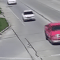} &
                                 \includegraphics[width=0.12\textwidth]{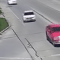 } &
                                 \includegraphics[width=0.12\textwidth]{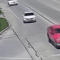} &
                                 \includegraphics[width=0.12\textwidth]{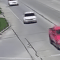} \\
\end{tabular}
  \caption{\textbf{Real-Traffic} Example 4}
  \label{rf-exmaple-4}
\end{figure*}
\begin{figure*}[t]
  \centering
  \small 
  \setlength{\tabcolsep}{1pt} 
  \begin{tabular}{c cc|ccccccc} 
    & \multicolumn{2}{c}{\centering{Input}} &   \multicolumn{7}{c}{\centering{Prediction}}   \\
    & \textbf{$t=1$} & \textbf{$t=5$} & \textbf{$t=7$} & \textbf{$t=9$} & \textbf{$t=11$} & \textbf{$t=13$} &\textbf{$t=15$} & \textbf{$t=18$} & \textbf{$t=25$}  \\
    \rotatebox{90}{GT}            & \includegraphics[width=0.1\textwidth]{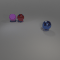} &
                                    \includegraphics[width=0.1\textwidth]{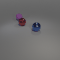} &
                                    \includegraphics[width=0.1\textwidth]{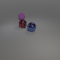 } &
                                    \includegraphics[width=0.1\textwidth]{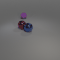 } &
                                    \includegraphics[width=0.1\textwidth]{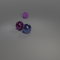 } &
                                    \includegraphics[width=0.1\textwidth]{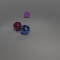} &
                                    \includegraphics[width=0.1\textwidth]{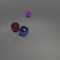} &
                                    \includegraphics[width=0.1\textwidth]{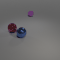} &
                                    \includegraphics[width=0.1\textwidth]{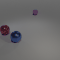} \\
\end{tabular}
  \caption{\textbf{CLEVR-3} Example 1}
  \label{clevr-3-exmaple-1}
\end{figure*}
\begin{figure*}[t]
  \centering
  \small 
  \setlength{\tabcolsep}{1pt} 
  \begin{tabular}{c cc|ccccccc} 
    & \multicolumn{2}{c}{\centering{Input}} &   \multicolumn{7}{c}{\centering{Prediction}}   \\
    & \textbf{$t=1$} & \textbf{$t=5$} & \textbf{$t=7$} & \textbf{$t=9$} & \textbf{$t=11$} & \textbf{$t=13$} &\textbf{$t=15$} & \textbf{$t=18$} & \textbf{$t=25$}  \\
    \rotatebox{90}{GT}            & \includegraphics[width=0.1\textwidth]{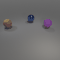} &
                                    \includegraphics[width=0.1\textwidth]{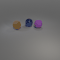} &
                                    \includegraphics[width=0.1\textwidth]{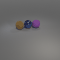 } &
                                    \includegraphics[width=0.1\textwidth]{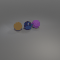 } &
                                    \includegraphics[width=0.1\textwidth]{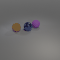 } &
                                    \includegraphics[width=0.1\textwidth]{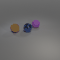} &
                                    \includegraphics[width=0.1\textwidth]{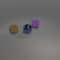} &
                                    \includegraphics[width=0.1\textwidth]{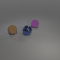} &
                                    \includegraphics[width=0.1\textwidth]{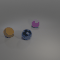} \\
\end{tabular}
  \caption{\textbf{CLEVR-3} Example 2}
  \label{clevr-3-exmaple-2}
\end{figure*}
\begin{figure*}[t]
  \centering
  \small 
  \setlength{\tabcolsep}{1pt} 
  \begin{tabular}{c cc|ccccccc} 
    & \multicolumn{2}{c}{\centering{Input}} &   \multicolumn{7}{c}{\centering{Prediction}}   \\
    & \textbf{$t=1$} & \textbf{$t=5$} & \textbf{$t=7$} & \textbf{$t=9$} & \textbf{$t=11$} & \textbf{$t=13$} &\textbf{$t=15$} & \textbf{$t=18$} & \textbf{$t=25$}  \\
    \rotatebox{90}{GT}            & \includegraphics[width=0.1\textwidth]{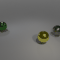} &
                                    \includegraphics[width=0.1\textwidth]{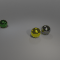} &
                                    \includegraphics[width=0.1\textwidth]{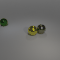 } &
                                    \includegraphics[width=0.1\textwidth]{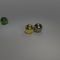 } &
                                    \includegraphics[width=0.1\textwidth]{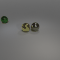 } &
                                    \includegraphics[width=0.1\textwidth]{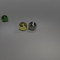} &
                                    \includegraphics[width=0.1\textwidth]{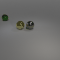} &
                                    \includegraphics[width=0.1\textwidth]{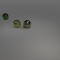} &
                                    \includegraphics[width=0.1\textwidth]{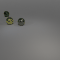} \\
\end{tabular}
  \caption{\textbf{CLEVR-3} Example 3}
  \label{clevr-3-exmaple-3}
\end{figure*}
\begin{figure*}[t]
  \centering
  \small 
  \setlength{\tabcolsep}{1pt} 
  \begin{tabular}{c cc|ccccccc} 
    & \multicolumn{2}{c}{\centering{Input}} &   \multicolumn{7}{c}{\centering{Prediction}}   \\
    & \textbf{$t=1$} & \textbf{$t=5$} & \textbf{$t=7$} & \textbf{$t=9$} & \textbf{$t=11$} & \textbf{$t=13$} &\textbf{$t=15$} & \textbf{$t=18$} & \textbf{$t=25$}  \\
    \rotatebox{90}{GT}            & \includegraphics[width=0.1\textwidth]{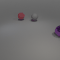} &
                                    \includegraphics[width=0.1\textwidth]{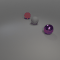} &
                                    \includegraphics[width=0.1\textwidth]{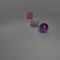 } &
                                    \includegraphics[width=0.1\textwidth]{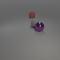 } &
                                    \includegraphics[width=0.1\textwidth]{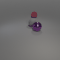 } &
                                    \includegraphics[width=0.1\textwidth]{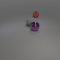} &
                                    \includegraphics[width=0.1\textwidth]{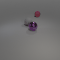} &
                                    \includegraphics[width=0.1\textwidth]{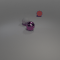} &
                                    \includegraphics[width=0.1\textwidth]{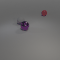} \\
\end{tabular}
  \caption{\textbf{CLEVR-3} Example 4}
  \label{clevr-3-exmaple-4}
\end{figure*}
\begin{figure*}[t]
  \centering
  \small 
  \setlength{\tabcolsep}{1pt} 
  \begin{tabular}{c cc|ccccccc} 
    & \multicolumn{2}{c}{\centering{Input}} &   \multicolumn{5}{c}{\centering{Prediction}}   \\
    & \textbf{$t=1$} & \textbf{$t=5$} & \textbf{$t=7$} & \textbf{$t=10$} & \textbf{$t=13$} & \textbf{$t=16$} &\textbf{$t=20$}  \\
    \rotatebox{90}{GT}            & \includegraphics[width=0.12\textwidth]{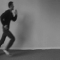} &
                                    \includegraphics[width=0.12\textwidth]{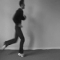} &
                                    \includegraphics[width=0.12\textwidth]{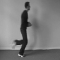 } &
                                    \includegraphics[width=0.12\textwidth]{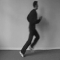 } &
                                    \includegraphics[width=0.12\textwidth]{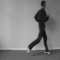 } &
                                    \includegraphics[width=0.12\textwidth]{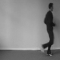} &
                                    \includegraphics[width=0.12\textwidth]{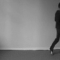} \\
     & \multicolumn{2}{c}{SiS}    &
                                    \includegraphics[width=0.12\textwidth]{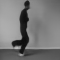 } &
                                    \includegraphics[width=0.12\textwidth]{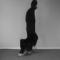 } &
                                    \includegraphics[width=0.12\textwidth]{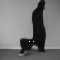 } &
                                    \includegraphics[width=0.12\textwidth]{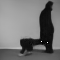} &
                                    \includegraphics[width=0.12\textwidth]{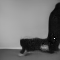} \\
\end{tabular}
  \caption{\textbf{KTH} Example 1}
  \label{Kubric-exmaple-2}
\end{figure*}
\begin{figure*}[t]
  \centering
  \small 
  \setlength{\tabcolsep}{1pt} 
  \begin{tabular}{c cc|ccccccc} 
    & \multicolumn{2}{c}{\centering{Input}} &   \multicolumn{5}{c}{\centering{Prediction}}   \\
    & \textbf{$t=1$} & \textbf{$t=5$} & \textbf{$t=7$} & \textbf{$t=10$} & \textbf{$t=13$} & \textbf{$t=16$} &\textbf{$t=20$}  \\
    \rotatebox{90}{GT}            & \includegraphics[width=0.12\textwidth]{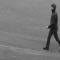} &
                                    \includegraphics[width=0.12\textwidth]{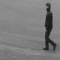} &
                                    \includegraphics[width=0.12\textwidth]{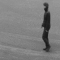 } &
                                    \includegraphics[width=0.12\textwidth]{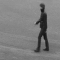 } &
                                    \includegraphics[width=0.12\textwidth]{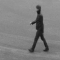 } &
                                    \includegraphics[width=0.12\textwidth]{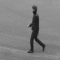} &
                                    \includegraphics[width=0.12\textwidth]{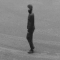} \\
     & \multicolumn{2}{c}{SiS}    &
                                    \includegraphics[width=0.12\textwidth]{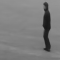 } &
                                    \includegraphics[width=0.12\textwidth]{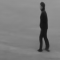 } &
                                    \includegraphics[width=0.12\textwidth]{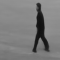 } &
                                    \includegraphics[width=0.12\textwidth]{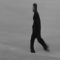} &
                                    \includegraphics[width=0.12\textwidth]{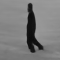} \\
\end{tabular}
  \caption{\textbf{KTH} Example 2}
  \label{kth-exmaple-2}
\end{figure*}
\begin{figure*}[t]
  \centering
  \small 
  \setlength{\tabcolsep}{1pt} 
  \begin{tabular}{c cc|ccccccc} 
    & \multicolumn{2}{c}{\centering{Input}} &   \multicolumn{5}{c}{\centering{Prediction}}   \\
    & \textbf{$t=1$} & \textbf{$t=5$} & \textbf{$t=7$} & \textbf{$t=10$} & \textbf{$t=13$} & \textbf{$t=16$} &\textbf{$t=20$}  \\
    \rotatebox{90}{GT}            & \includegraphics[width=0.12\textwidth]{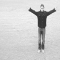} &
                                    \includegraphics[width=0.12\textwidth]{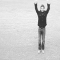} &
                                    \includegraphics[width=0.12\textwidth]{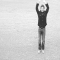 } &
                                    \includegraphics[width=0.12\textwidth]{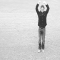 } &
                                    \includegraphics[width=0.12\textwidth]{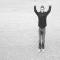 } &
                                    \includegraphics[width=0.12\textwidth]{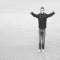} &
                                    \includegraphics[width=0.12\textwidth]{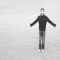} \\
     & \multicolumn{2}{c}{SiS}    &
                                    \includegraphics[width=0.12\textwidth]{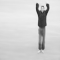 } &
                                    \includegraphics[width=0.12\textwidth]{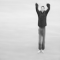 } &
                                    \includegraphics[width=0.12\textwidth]{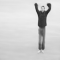 } &
                                    \includegraphics[width=0.12\textwidth]{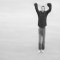} &
                                    \includegraphics[width=0.12\textwidth]{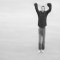} \\
\end{tabular}
  \caption{\textbf{KTH} Example 3}
  \label{kth-exmaple-3}
\end{figure*}
\begin{figure*}[t]
  \centering
  \small 
  \setlength{\tabcolsep}{1pt} 
  \begin{tabular}{c cc|ccccccc} 
    & \multicolumn{2}{c}{\centering{Input}} &   \multicolumn{5}{c}{\centering{Prediction}}   \\
    & \textbf{$t=1$} & \textbf{$t=5$} & \textbf{$t=7$} & \textbf{$t=10$} & \textbf{$t=13$} & \textbf{$t=16$} &\textbf{$t=20$}  \\
    \rotatebox{90}{GT}            & \includegraphics[width=0.12\textwidth]{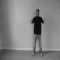} &
                                    \includegraphics[width=0.12\textwidth]{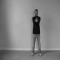} &
                                    \includegraphics[width=0.12\textwidth]{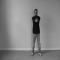 } &
                                    \includegraphics[width=0.12\textwidth]{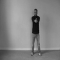 } &
                                    \includegraphics[width=0.12\textwidth]{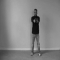 } &
                                    \includegraphics[width=0.12\textwidth]{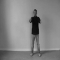} &
                                    \includegraphics[width=0.12\textwidth]{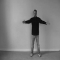} \\
     & \multicolumn{2}{c}{SiS}    &
                                    \includegraphics[width=0.12\textwidth]{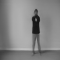 } &
                                    \includegraphics[width=0.12\textwidth]{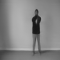 } &
                                    \includegraphics[width=0.12\textwidth]{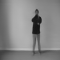 } &
                                    \includegraphics[width=0.12\textwidth]{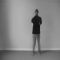} &
                                    \includegraphics[width=0.12\textwidth]{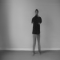} \\
\end{tabular}
  \caption{\textbf{KTH} Example 4}
  \label{kth-exmaple-4}
\end{figure*}
\begin{figure*}[t]
  \centering
  \small 
  \setlength{\tabcolsep}{1pt} 
  \begin{tabular}{c cc|ccccccc} 
    & \multicolumn{2}{c}{\centering{Input}} &   \multicolumn{5}{c}{\centering{Prediction}}   \\
    & \textbf{$t=1$} & \textbf{$t=5$} & \textbf{$t=7$} & \textbf{$t=10$} & \textbf{$t=13$} & \textbf{$t=16$} &\textbf{$t=20$}  \\
    \rotatebox{90}{GT}            & \includegraphics[width=0.12\textwidth]{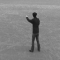} &
                                    \includegraphics[width=0.12\textwidth]{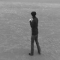} &
                                    \includegraphics[width=0.12\textwidth]{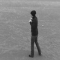 } &
                                    \includegraphics[width=0.12\textwidth]{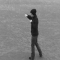 } &
                                    \includegraphics[width=0.12\textwidth]{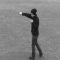 } &
                                    \includegraphics[width=0.12\textwidth]{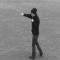} &
                                    \includegraphics[width=0.12\textwidth]{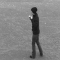} \\
     & \multicolumn{2}{c}{SiS}    &
                                    \includegraphics[width=0.12\textwidth]{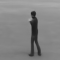 } &
                                    \includegraphics[width=0.12\textwidth]{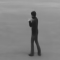 } &
                                    \includegraphics[width=0.12\textwidth]{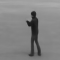 } &
                                    \includegraphics[width=0.12\textwidth]{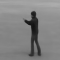} &
                                    \includegraphics[width=0.12\textwidth]{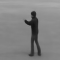} \\
\end{tabular}
  \caption{\textbf{KTH} Example 5}
  \label{kth-exmaple-5}
\end{figure*}

\end{document}